\documentclass[a4paper,11pt]{article}

\usepackage[a4paper, left=3cm, right=3cm]{geometry}

\usepackage{tango-colors}

\usepackage{tikz}
\usetikzlibrary{arrows}
\usetikzlibrary{shapes}
\usetikzlibrary{shapes.misc}

\usepackage{pgfplots}
\pgfplotsset{compat=1.9}
\usepgfplotslibrary{groupplots}
\usepgfplotslibrary{polar}

\usepackage{array}
\usepackage{booktabs}

\newcolumntype{C}{>{\centering\arraybackslash}p{1.7cm}}
\newcolumntype{W}{>{\centering\arraybackslash}p{2cm}}

\usepackage{algorithm}
\usepackage{algpseudocode}

\usepackage{bm}
\usepackage{amssymb}
\usepackage{amsthm}
\usepackage{mathtools}

\usepackage{authblk}

\usepackage[utf8]{inputenc}
\usepackage[T1]{fontenc}
\usepackage[english]{babel}
\usepackage{csquotes}
\usepackage{hyperref}

\newtheorem{thm}{Theorem}
\theoremstyle{definition}

\newcommand{\R}{\mathbb{R}}
\newcommand{\N}{\mathbb{N}}
\newcommand{\transp}[1]{{#1}^T}

\newcommand{\grad}[1]{\nabla_{#1}}

\newcommand{\hess}[1]{\nabla^2_{#1}}
\newcommand{\pInv}[1]{{#1}^\dag}
\DeclareMathOperator*{\argmin}{argmin}
\newcommand{\eye}{I}
\newcommand{\effic}{\mathcal{O}}

\newcommand{\prob}{P}
\newcommand{\dens}{p}
\newcommand{\expec}{\mathbb{E}}

\newcommand{\nDens}{\mathcal{N}}
\newcommand{\nMean}{\vec \mu}
\newcommand{\nPrec}{\Lambda}
\newcommand{\nDev}{\sigma}

\newcommand{\datum}{\vec x}
\newcommand{\data}{X}
\newcommand{\dataidx}{i}
\newcommand{\datalim}{M}

\newcommand{\lbl}{y}

\newcommand{\lbllim}{L}

\newcommand{\nonlin}{\Phi}

\newcommand{\protolbl}{\bar \lbl}
\newcommand{\protos}{W}
\newcommand{\protoidx}{k}
\newcommand{\protolim}{K}
\newcommand{\relmat}{\Omega}

\newcommand{\dist}{d}

\newcommand{\model}{f}

\newcommand{\srcspace}{\mathcal{X}}
\newcommand{\dimsrc}{m}
\newcommand{\tarspace}{\hat \srcspace}
\newcommand{\tardata}{\hat X}
\newcommand{\tardatum}{\hat x}
\newcommand{\tardataidx}{j}
\newcommand{\tardatalim}{N}
\newcommand{\tarlbl}{\hat y}
\newcommand{\dimtar}{n}
\newcommand{\tartosrc}{h}
\newcommand{\transf}{H}
\newcommand{\err}{E}
\newcommand{\regul}{\lambda}

\newcommand{\pstr}{\gamma}
\newcommand{\pstrmat}{\Gamma}
\newcommand{\qerr}{E_Q}

\newcommand{\fig}{figure}
\newcommand{\tab}{table}

\tikzstyle{point}=[circle, inner sep=0pt, minimum size=5mm, line width=0.5mm, anchor=center]
\tikzstyle{textnode}=[draw=none, fill=none]
\tikzstyle{proto}=[diamond, inner sep=0pt, minimum size=7mm, line width=0.5mm, anchor=center]
\tikzstyle{edge}=[->, >=stealth', shorten <=2pt, shorten >=2pt, auto, line width=0.5mm]
\tikzstyle{class0color}=[aluminium6]
\tikzstyle{class0}=[draw=aluminium6, fill=aluminium4, text=aluminium6]
\tikzstyle{class1color}=[skyblue3]
\tikzstyle{class1}=[draw=skyblue3, fill=skyblue1, text=skyblue3]
\tikzstyle{class2color}=[orange3]
\tikzstyle{class2}=[draw=orange3, fill=orange1, text=orange3]

\begin{document}

\title{Expectation maximization transfer learning and its application for bionic hand prostheses}

\author[1]{Benjamin Paaßen}
\author[1]{Alexander Schulz}
\author[2]{Janne Hahne}
\author[1]{Barbara Hammer}

\affil[1]{Center of Excellence Cognitive Interaction Technology\protect\\
Bielefeld University\protect\\
\{bpaassen|aschulz|bhammer\}@techfak.uni-bielefeld.de}

\affil[2]{Neurorehabilitation Systems Research Group\protect\\
Department of Trauma Surgery\protect\\
Orthopedic Surgery and Hand Surgery\protect\\
Universiy Medical Center Göttingen}

\date{This is a preprint of a publication to appear in the Journal \emph{Neurocomputing} (Special Issue regarding ESANN 2017)
as provided by the authors.}

\pagestyle{myheadings}
\markright{Preprint as provided by the authors.}

\maketitle

\begin{abstract}
Machine learning models in practical settings are typically confronted with
changes to the distribution of the incoming data.
Such changes can severely affect the model performance, leading for example to
misclassifications of data. This is particularly apparent in the domain of bionic hand prostheses,
where machine learning models promise faster and more intuitive user interfaces, but are hindered
by their lack of robustness to everyday disturbances, such as electrode shifts.
One way to address changes in the data distribution is transfer learning, that is,
to transfer the disturbed data to a space where the original model is applicable again.
In this contribution, we propose a novel expectation maximization algorithm to learn linear
transformations that maximize the likelihood of disturbed data according to the undisturbed model.
We also show that this approach generalizes to discriminative models, in particular learning
vector quantization models.
In our evaluation on data from the bionic prostheses domain we demonstrate that our approach can
learn a transformation which improves classification accuracy significantly and outperforms all
tested baselines, if few data or few classes are available in the target domain.
\end{abstract}

\section{Introduction}

Classical machine learning theory relies on the assumption that training and test data stem from
the same underlying distribution; an assumption, that is oftentimes violated in practical applications
\cite{Cortes2008}. The reasons for such violations are multifold. The training data may be selected
in a biased way and not represent the \enquote{true} distribution properly \cite{Cortes2008},
disturbances may lead to changes in the data over time \cite{Ditzler2015}, or one may try to transfer
an existing model to a new domain \cite{BenDavid2006}.
If such violations occur, the model may not accurately describe the data anymore, leading to errors,
e.g.\ in classification.

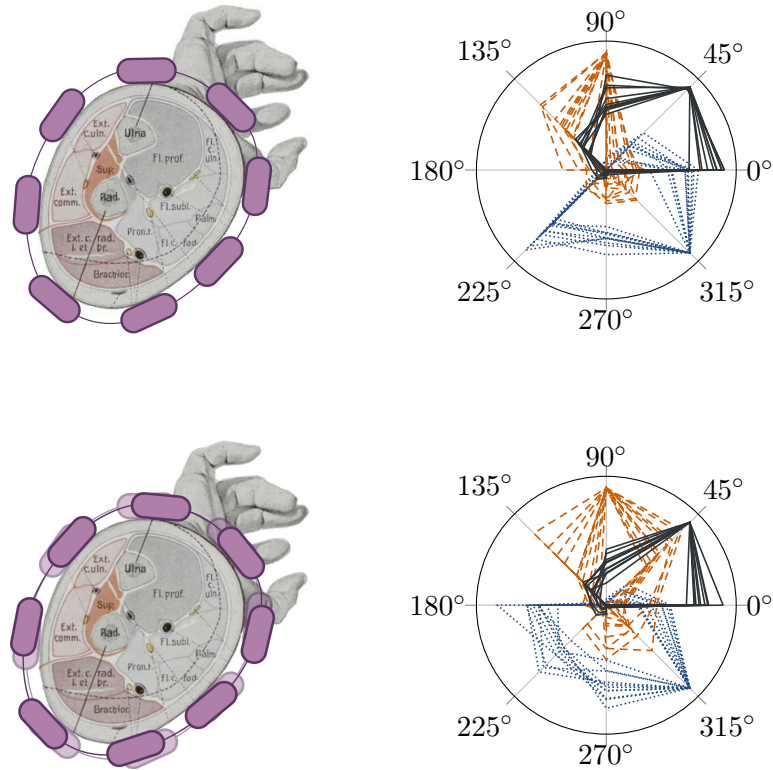
\begin{figure}
\begin{center}
\begin{tikzpicture}
\matrix[column sep=1cm, row sep=1cm]{
\node at (0, 1.8)  {\includegraphics[width=4cm]{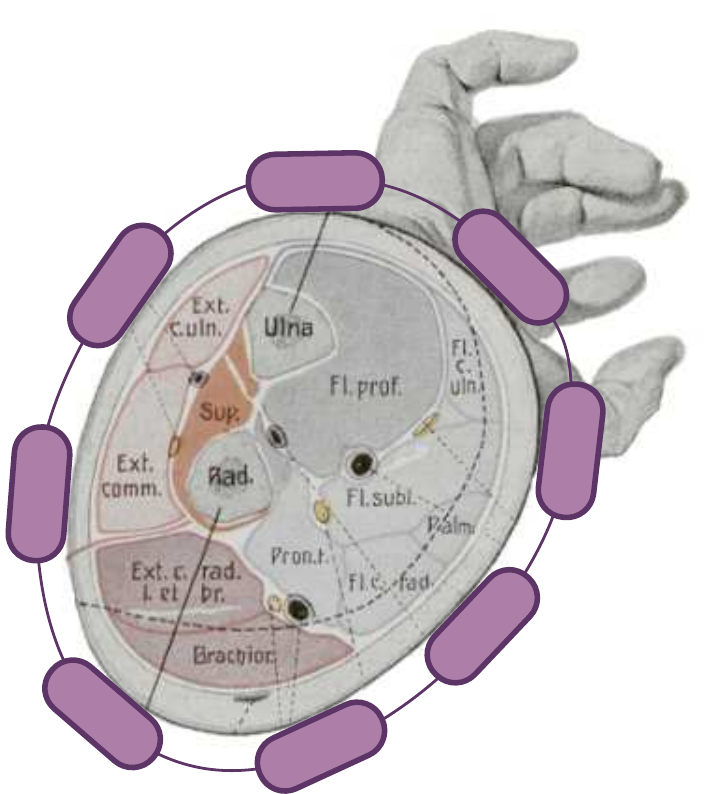}};
&
\begin{polaraxis}[
	no markers,
	xticklabel=$\pgfmathprintnumber{\tick}^\circ$,
	xtick={0,45,...,315},
	hide y axis,
	width=5cm,
]
\foreach \i in {1,...,11}{
	\addplot[class1color, semithick, densely dotted]%
	table[data cs=polarrad, x=angles, y=point_1_\i] {emg_example_data.csv};
}
\foreach \i in {1,...,11}{
	\addplot[class2color, semithick, densely dashed]%
	table[data cs=polarrad, x=angles, y=point_2_\i] {emg_example_data.csv};
}
\foreach \i in {1,...,11}{
	\addplot[class0color, semithick]%
	table[data cs=polarrad, x=angles, y=point_3_\i] {emg_example_data.csv};
}
\end{polaraxis}
\\
\node at (0, 1.8) {\includegraphics[width=4cm]{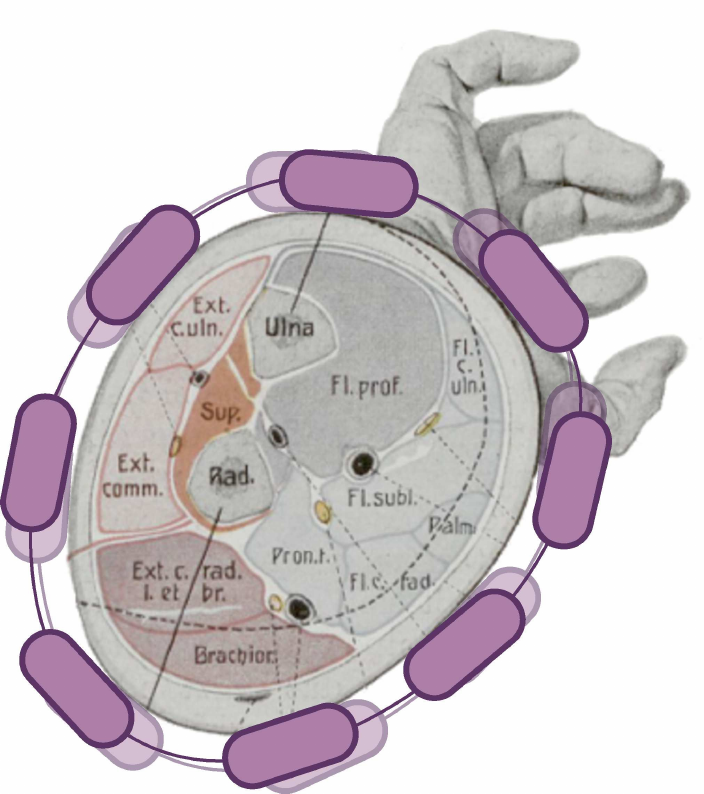}};
&
\begin{polaraxis}[
	no markers,
	xticklabel=$\pgfmathprintnumber{\tick}^\circ$,
	xtick={0,45,...,315},
	hide y axis,
	width=5cm,
]
\foreach \i in {1,...,11}{
	\addplot[class1color, semithick, densely dotted]%
	table[data cs=polarrad, x=angles, y=point_1_\i] {emg_example_data_shifted.csv};
}
\foreach \i in {1,...,11}{
	\addplot[class2color, semithick, densely dashed]%
	table[data cs=polarrad, x=angles, y=point_2_\i] {emg_example_data_shifted.csv};
}
\foreach \i in {1,...,11}{
	\addplot[class0color, semithick]%
	table[data cs=polarrad, x=angles, y=point_3_\i] {emg_example_data_shifted.csv};
}
\end{polaraxis}
\\
};
\end{tikzpicture}
\end{center}
\caption{An illustration of electrode shifts in electromyographic (EMG) data.
Top left: A grid of eight EMG electrodes placed around the forearm of a user.
Cross section of the arm taken from the
1921 German edition of \enquote{Anatomie des Menschen}, which is in the public domain.
Top right: Example EMG signals from an eight-electrode EMG recording for two different
hand motions (dashed and dotted lines) as well as resting (solid lines).
Bottom left: The electrode grid is shifted around the forearm (electrode shift).
Bottom right: Another set of of EMG signals from a shifted eight-electrode EMG recording
for two different hand motions (dashed and dotted lines) as well as resting (solid lines).
Due to the shifted signal, a model trained on the source data (top right) may misclassify
shifted data (bottom right).}
\label{fig:electrode_shift}
\end{figure}

This is particularly apparent in the domain of bionic hand prostheses. By now, research prototypes
of such prostheses feature up to 20 active degrees of freedom (DoF), promising to restore precise and
differentiated hand functions \cite{Belter2013}. However, controlling this many degrees of freedom
requires a user interface which reacts rapidly and is intuitive to the user. A popular approach to
achieve such a user interface is to let users execute the desired motion with their phantom hand,
which is still represented in the brain, and infer the desired motion via classification of the
residual muscle signals in the forearm, such that the desired motion can then be executed by
a bionic hand prosthesis in real-time (time delay below 200 ms) \cite{Farina2014}. More precisely,
if a user executes a motion with her phantom hand, the corresponding neurons in the brain
are activated and propagate the motor command to the arm, where the residual muscles
responsible for the hand motion are activated. This activity can be recorded via a grid of
electromyographic (EMG) electrodes placed on the skin around the amputee's forearm
(see \fig\ \ref{fig:electrode_shift}, top left). The EMG signal contains information about the
firing pattern of the motor neurons, which in turn codes the intended hand motion.
Therefore, one can classify the EMG signal with respect to the intended hand motion and use the
classification result to control a prosthesis with little time delay in an intuitive way \cite{Farina2014}.

Unfortunately, such user interfaces are seriously challenged by changes in the input data
distribution due to disturbances to the EMG signal, for example by electrode shifts,
posture changes, sweat, fatigue, etc. \cite{Farina2014,Khushaba2014}. As an example, consider
\fig\ \ref{fig:electrode_shift}, which illustrates the effect of an electrode shift around the
forearm, leading to different EMG sensor data, which in turn may cause an erroneous classification
decision.

Changes between training and test distribution have been addressed by different theoretical
frameworks. Shimodaira has introduced the notion of \emph{covariate shift} describing the case
of a change in the prior distribution $\dens(\datum)$ while the conditional distribution of the
label $\prob(\lbl | \datum)$ remains unchanged \cite{Shimodaira2000}. A slightly different
angle is taken by \emph{sample selection bias correction theory} which assumes that a \emph{true}
underlying distribution $\prob(\lbl, \datum)$ exists from which some pairs are not available
in the training data, thereby biasing the resulting machine learning model \cite{Cortes2008}.
In contrast, the theory of \emph{concept drift} models the prior distribution
$\dens(\datum)$ and the conditional distribution $\prob(\lbl | \datum)$ as varying
in time. In particular, a covariate shift, meaning a change in $\dens(\datum)$ over
time while $\prob(\lbl | \datum)$ stays constant, is called \emph{virtual concept drift}.
A change in $\prob(\lbl | \datum)$ over time is called \emph{real concept drift}.
Prior research in concept drift has focussed on either adapting a model over time to smooth and
slow concept drifts or detecting a point of sudden concept drift, such that the old model can be
discarded and a new model can be learned \cite{Ditzler2015}. Recently, explicit long
and short term memory models demonstrated an excellent ability to cope with different types of
concept drift \cite{Losing2016}.

Our example of electrode shifts in bionic hand prostheses is best described by a sudden, real
concept drift, in which case concept drift theory would recommend to discard the existing classifier
and re-train a new one \cite{Ditzler2015}. However, re-learning a viable classifier model may
require considerable amounts of new training data to be recorded, which is inconvenient or even
infeasible in user's everyday lives. Instead, we would like to re-use an existing classifier
model and \emph{adapt} it to the disturbed situation. This approach is motivated by prior
research on myoelectric data which indicates that disturbances to electrode shifts are typically
simple in structure, that is, they tend to be signal amplitude changes and shifts in the frequency
spectrum \cite{Khushaba2014}. Therefore, learning to transfer between the disturbed and the
undisturbed setting may be considerably simpler compared to learning a new model \cite{Paassen2016NC2}.

Learning such transfers between domains has been studied in the fields of \emph{domain adaptation}
and \emph{transfer learning}. Domain adaptation refers to re-using an existing model in another
domain where little to none new training samples are available \cite{BenDavid2006}. Similarly,
transfer learning refers to the transfer of knowledge from a source domain, where a viable model
is available, to a target domain, where the prior and/or conditional distribution is different \cite{Pan2010}.
In particular, rather than adjusting the probability distribution in a given data space, transfer
learning focusses on adapting the data \emph{representation}. Conceptually, this fits well to
our setting as the data representation in terms of EMG readings changes, while the underlying data
source, i.e.\ the neural code of the desired motion, remains the same.

Our key contribution is an efficient algorithm for transfer learning on labeled Gaussian mixture
models relying on expectation maximization \cite{Dempster1977}.
In particular, we learn a linear transformation which maps the target space training data
to the source space such that the likelihood of the target space data according to the source space model is
maximized. This approach generalizes to discriminative models, in particular learning vector
quantization models, such as generalized matrix learning vector quantization (GMLVQ), or its
localized version, LGMLVQ \cite{Schneider2009}.
We evaluate our approach on artificial as well as real myoelectric data and show that our transfer
learning approach can learn a transfer mapping which improves classification accuracy significantly
and outperforms all tested baselines, if few samples from the target space are available and/or these
samples do not cover all classes.

We begin by discussing related work, continue by introducing our own approach and conclude
by evaluating our approach in comparisons to baselines from the literature.

\section{Related Work\label{sec:rel_work}}

We begin our comparison to related work by introducing some key concepts of transfer learning
more formally. In our setting, we assume that a classification model $\model : \srcspace \to \{1, \ldots, \lbllim\}$
has been trained in some source space $\srcspace = \R^\dimsrc$ for some $\dimsrc \in \N$ and we want
to apply this model $\model$ in some target space $\tarspace = \R^\dimtar$ for some $\dimtar \in \N$.
Note that we assume that the classification task itself is the same for both spaces.
This makes our setup an instance of \emph{domain adaptation} \cite{BenDavid2006} or
\emph{transductive transfer learning} \cite{Pan2010}.
In the example of an electrode shift on EMG data, we have $\dimsrc = \dimtar$, but a simple
application of our source space classifier $\model$ is hindered by a fact that the activation
pattern is rotated in the feature space and thus the joint distribution $\dens_{\tarspace}(\tardatum, \tarlbl)$
for data $\tardatum \in \tarspace$ and labels $\tarlbl \in \{1, \ldots, \lbllim\}$ in the target space
differs from the joint distribution in the source space $\dens_{\srcspace}(\datum, \lbl)$
(see \fig\ \ref{fig:electrode_shift}).

One family of approaches to address domain adaptation problems are importance sampling approaches,
such as kernel mean matching \cite{Huang2006NIPS}, which apply a weight to each data point in the source
space and re-learn the model $\model$ with these weighted data points in order to generalize better
to the target space \cite{Pan2010}. The weights approximate the fraction
$\frac{\dens_{\tarspace}(\datum)}{\dens_{\srcspace}(\datum)}$, that is,
the proportion of the probability of a point in the source space and in the target space.
It can be shown that these weights minimize the empirical risk in the target space, if the
conditional distributions in both spaces are equal, that is,
$\dens_{\tarspace}(\lbl | \datum) = \dens_{\srcspace}(\lbl | \datum)$ \cite{Pan2010}. However,
this rather demanding assumption does not hold in our case because electrodes shift on EMG data
also influence the distribution of labels.

\begin{figure}
\begin{center}
\begin{tikzpicture}
\begin{groupplot}[
	disabledatascaling,
	group style={
		group size=2 by 1, horizontal sep=0.5cm, vertical sep=0.8cm,
		x descriptions at=edge bottom,
		y descriptions at=edge left,
	},
	width = 0.525\textwidth,
	scatter/classes={%
		1={mark=*,class1, mark size=0.5mm, opacity=0.6},%
		2={mark=square*,class0, mark size=0.5mm, opacity=0.6},%
		3={mark=diamond*,class2, mark size=0.8mm, opacity=0.6}%
	},%
	enlarge x limits=0.3,
	enlarge y limits,
]
\nextgroupplot[title = {$\data$}, axis equal, xmin=-2, xmax=2, ymin=-2, ymax=2]
\addplot[scatter,only marks,
scatter src=explicit symbolic]
table[x=x, y=y, meta=label] {toy_data_src.csv};
\draw[densely dotted, class1, fill=none, thick]  (-1, 0) circle (0.6);
\draw[densely dotted, class0, fill=none, thick]  (0, 0) circle (0.6);
\draw[densely dotted, class2, fill=none, thick]  (1, 0) circle (0.6);
\nextgroupplot[title = {$\tardata$}, axis equal, xmin=-2, xmax=2, ymin=-2, ymax=2]
\addplot[scatter,only marks,
scatter src=explicit symbolic]
table[x=x, y=y, meta=label] {toy_data_tar.csv};
\begin{scope}[rotate={-90},scale={2}]
\draw[densely dotted, class1, fill=none, thick]  (-1, 0.05) circle (0.3);
\draw[densely dotted, class0, fill=none, thick]  (0, 0) circle (0.3);
\draw[densely dotted, class2, fill=none, thick]  (1, -0.05) circle (0.3);
\end{scope}
\end{groupplot}
\end{tikzpicture}
\end{center}
\caption{A simple two-dimensional data set illustrating transfer learning for a
classification problem with three classes.
The source space data (left) is generated by a labeled Gaussian mixture model with one component
for the first (circles), the second (squares), and the third class (diamonds) respectively.
The Gaussians have means $(-1, 0)$, $(0, 0)$ and $(1, 0)$ and standard deviation $0.3$
(indicated by dashed circles at two standard deviations).
The target space data (right) is generated by a similar model, but the Gaussian means
are now at $(-0.1, -2)$, $(0, 0)$ and $(0.1, 2)$ respectively.}
\label{fig:toy_data}
\end{figure}
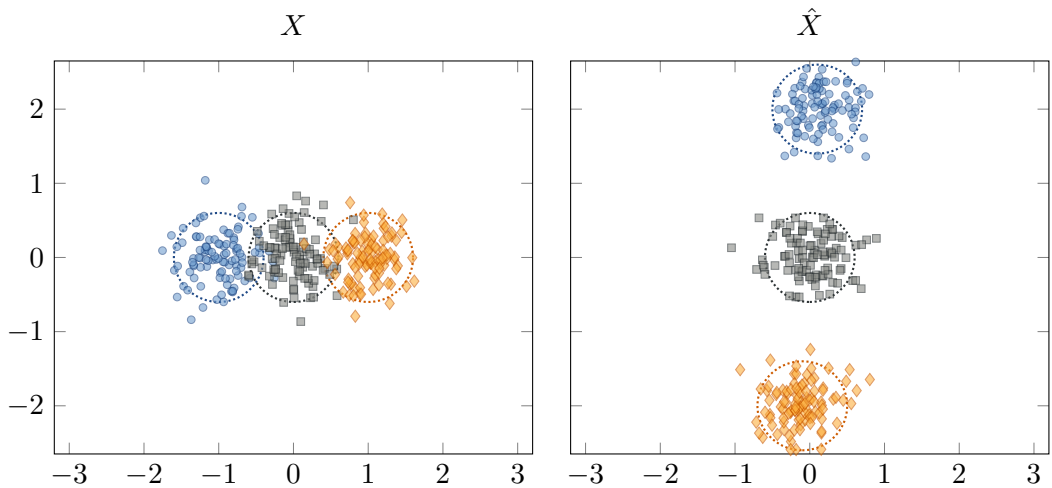

Another family of approaches attempts to map the data from both the source and the target space
to a shared latent space in which a model can be learned that applies to both the source and
the target space \cite{Pan2010,Bloebaum2015}. While these approaches are more general, they typically do
not take label information into account, which makes transfer learning in cases such as ours
significantly more challenging.
Consider the simple example in \fig\ \ref{fig:toy_data}, which displays a data set with
three classes, where the class-specific distribution $\dens(\datum | \lbl)$ for each
label $\lbl$ is given as a Gaussian with means means $(-1, 0)$, $(0, 0)$, and $(1, 0)$ respectively.
A classifier trained on this data set is likely to identify the $x_1$-axis as discriminative
dimension and assign every data point with $x_1 \leq -0.5$ to class 1, every data point with
$-0.5 < x_1 < 0.5$ to class 2 and any other data point to class 3.
Now, we want to transfer this model to the target space where the data representation has
changed (namely, the data set is rotated and the means are moved further apart).
Without any information regarding the conditional distribution of the target space data,
it is inherently difficult to learn a viable transfer mapping because the assignment between
the clusters in the target space and the clusters in the source
space is ambiguous. In particular, a solution that maximizes the correspondence of the marginal
distributions is to map the mean of the class 1 cluster to the mean of the class 3 cluster
and vice versa. Conversely, if label information from the target space is available, we
can disambiguate the assignment and thus simplify the transfer learning problem.

One approach which does take label information into account is the adaptive support vector
machine (a-SVM) \cite{Yang2007}. It assumes that some classifier
has been trained on source space data and can assign labels to the labeled target space
training data points. The a-SVM then attempts to predict the difference between the
predicted labels by the source space classifier and the actual target space labels, such that
it does not have to classify points which are already correctly classified by the original
classifier \cite{Yang2007}. Note that for our example in \fig\ \ref{fig:toy_data}, the source
model classification is only correct for the middle cluster and the a-SVM needs to re-learn
the classification for the remaining two classes even though the change in representation
between source and target space is structurally simple. In that sense, the a-SVM does more
than is necessary in this case and does not exploit our knowledge about the task perfectly.
Still, we will consider it as a baseline model in our experiments.

A recent framework which tries to explicitly learn the change in representation between source
and target space is linear supervised transfer learning \cite{Paassen2016NC2}. Assuming some
classifier $\model : \srcspace \to \{1, \ldots, \lbllim\}$ in the source space, the approach
attempts to learn a mapping $\tartosrc : \tarspace \to \srcspace$
which minimizes the error of the classifier $\model \circ \tartosrc$ on the labeled target space
data. In other words, the approach attempts to learn the change in representation between target
and source space such that the target space data can be classified correctly by the source space
classifier after mapping the data to the source space via $\tartosrc$ \cite{Paassen2016NC2}.
Intuitively, such an approach is particularly promising if the relationship $\tartosrc$ is easier to
learn compared to a new classifier $\hat \model$ for the target space. To ensure this constraint,
the authors assume that $\tartosrc$ can be approximated by a linear function, that is, the
first-order Taylor approximation without constant term is a good approximation of $\tartosrc$ for
the given data \cite{Saralajew2017}. Note that the mapping $\tartosrc$ needs to be optimized with
respect to the model error, such that the optimization process is inherently specific to a certain
source space classifier. Until now, a gradient descent scheme on the cost function of
generalized learning matrix vector quantization (GMLVQ) has been suggested and shown to be effective
for EMG data classification \cite{Prahm2016}. In our contribution, we extend this work in
several key points. First, we provide a more precise notion of linear supervised transfer
learning in a probabilistic sense. Second, we provide a general expectation maximization
algorithm to optimize the model fit for labeled Gaussian mixture models. Third, we apply this
algorithm to models of the learning vector quantization family, such as GMLVQ and localized GMLVQ
\cite{Schneider2009}, and show that we can outperform prior approaches in terms of classification
accuracy on EMG data.

\section{Transfer Learning for Labeled Gaussian Mixtures}

We begin the description of our proposed approach by re-phrasing the basic problem of transfer
learning in probabilistic terms. We start with a source space $\srcspace$, in which labeled data
points $(\datum, \lbl)$ with $\lbl \in \{1, \ldots, \lbllim\}$ are generated according to a joint
probability density $\dens_{\srcspace}(\datum, \lbl)$. In this space, we train a source classifier
$\model : \srcspace \to \{1, \ldots, \lbllim\}$. Then, we wish to apply this classifier to data
in a target space $\tarspace = \R^\dimtar$ for some $\dimtar \in \N$, where labeled
data points $(\tardatum, \tarlbl)$ are generated according to the probability density
$\dens_{\tarspace}(\tardatum, \tarlbl)$ with $\tarlbl \in \{1, \ldots, \lbllim\}$.
Our basic assumption is that there exists a smooth function $\tartosrc : \tarspace \to \srcspace$
such that for all $\tardatum \in \tarspace$ and all $\tarlbl \in \{1, \ldots, \lbllim\}$ it holds:
$\dens_{\tarspace}(\tardatum, \tarlbl) = \dens_{\srcspace}(\tartosrc(\tardatum), \tarlbl)$.
In other words: We assume, that there is a smooth mapping $\tartosrc$ that characterizes the
\emph{change in representation} between target and source space and therefore fully explains the
differences in the joint probability densities.
We propose to learn this mapping $\tartosrc$ using a maximum likelihood approach. In particular, we
intend to construct a model of the joint probability density for data and labels
in the source space $\dens(\datum, \lbl)$. Then, we intend to maximize the empiric likelihood of
a labeled example data set from the target space 
$\{ (\tardatum_\tardataidx, \tarlbl_\tardataidx) \}_{\tardataidx \in \{1, \ldots, \tardatalim\}}$,
according to the source space model, that is, we attempt to solve the optimization problem
\begin{equation}
\max_{\tartosrc} \prod_{\tardataidx=1}^\tardatalim \dens\Big(\tartosrc(\tardatum_\tardataidx), \tarlbl_\tardataidx\Big)
\label{eq:tl}
\end{equation}

Note that, if we only care about classification, we do not require $\dens(\datum, \lbl)$ to be a
precise, generative model for the source space data. In this case, it suffices if we have a
precise model of $\prob(\lbl | \datum)$ as provided by many classifiers, whereas the
approximation of the data density $\dens(\datum)$ may be inaccurate. Therefore, we
disregard the specific difficulties of inferring a precise density model within this contribution.

In the remainder of this work, we will provide a solution for the maximization problem~\ref{eq:tl}
for a special class of models, namely Gaussian mixture models for $\dens(\datum, \lbl)$ and linear
functions $\tartosrc$.

\subsection{Labeled Gaussian Mixture Models}

First, we choose to approximate $\dens_\srcspace(\datum, \lbl)$ with a Gaussian mixture
model (GMM). Such models are well-established in machine learning \cite{Bishop2006,Barber2012}
and have been successfully applied to classify EMG data \cite{Huang2005}. In general, almost
any density can be approximated via a GMM, if a sufficient number of Gaussian components
$\protolim$ is employed \cite{Bishop2006}.
GMMs approximate a density $\dens(\datum)$ via a sum of Gaussians
$\protoidx = 1, \ldots, \protolim$ as follows \cite{Bishop2006,Barber2012}:
\begin{align}
\dens(\datum) &\approx \sum_{\protoidx=1}^\protolim \nDens(\datum|\nMean_\protoidx, \nPrec_\protoidx)
	\cdot \prob(\protoidx) & \text{where} \\
\nDens(\datum | \nMean_\protoidx, \nPrec_\protoidx) &=
	\sqrt{\frac{\det(\nPrec_\protoidx)}{(2 \cdot \pi)^\dimsrc}} \cdot
	\exp\Big(-\frac{1}{2} \cdot
		\transp{(\datum - \nMean_\protoidx)} \cdot \nPrec_\protoidx \cdot (\datum - \nMean_\protoidx)
	\Big) \label{eq:Gaussian}
\end{align}
where $\nMean_\protoidx$ is the mean of the $\protoidx$th Gaussian and $\nPrec_\protoidx$ is
the precision matrix (a positive definite matrix, which is the inverse of the covariance
matrix) of the $\protoidx$th Gaussian.

In our setting, we intend to apply Gaussian mixture models for classification, which means that
we need to include the label $\lbl$ of the data in the model. In particular, we re-write the
joint probability density $\dens(\datum, \lbl)$ as follows:
\begin{equation}
\dens(\datum, \lbl) = \sum_{\protoidx=1}^\protolim \dens(\datum, \lbl, \protoidx)
= \sum_{\protoidx=1}^\protolim
\dens(\datum | \lbl, \protoidx) \cdot \prob(\lbl | \protoidx) \cdot \prob(\protoidx)
\end{equation}
As before we assume that $\dens(\datum|\protoidx)$ is a Gaussian density. Additionally,
we assume that the label $\lbl$ of a data point and the data point $\datum$ itself are
conditionally independent given the Gaussian component $\protoidx$ they have been
generated from. Under these assumptions we obtain:
\begin{equation}
\dens(\datum, \lbl) = \sum_{\protoidx=1}^\protolim \nDens(\datum | \nMean_\protoidx, \nPrec_\protoidx)
	\cdot \prob(\lbl | \protoidx) \cdot \prob(\protoidx) \label{eq:lgmm}
\end{equation}
The parameters of our model are, for each Gaussian component $\protoidx$,
a mean $\nMean_\protoidx$, a precision matrix $\nPrec_\protoidx$, a probability distribution
over the labels $\prob(\lbl | \protoidx)$ and a prior probability $\prob(\protoidx)$.
A simple example of such a model is shown in \fig\ \ref{fig:toy_data} (left).
The parameters for this labeled Gaussian mixture model (lGMM) are $\nMean_1 = (-1, 0)$, $\nMean_2 = (0, 0)$, and
$\nMean_3 = (1, 0)$ as means, $\nPrec_1 = \nPrec_2 = \nPrec_3 = \frac{1}{0.3^2} \cdot \eye^\dimsrc$
as precision matrices, where $\eye^\dimsrc$ is the $\dimsrc \times \dimsrc$-dimensional identity matrix,
and $\prob(\lbl | \protoidx) = 1$ if $\lbl = \protoidx$ and $0$ otherwise, as well as
$\prob(\protoidx) = \frac{1}{3}$ for all $\protoidx$.

We can classify data with an lGMM via a maximum a posteriori approach, where the posterior is:
\begin{equation}
\prob(\lbl | \datum) = \frac{\dens(\datum, \lbl)}{\dens(\datum)}
= \frac{
	\sum_{\protoidx = 1}^\protolim  \nDens(\datum | \nMean_\protoidx, \nPrec_\protoidx) \cdot \prob(\lbl | \protoidx) \cdot \prob(\protoidx)
}{
	\sum_{\protoidx = 1}^\protolim  \nDens(\datum | \nMean_\protoidx, \nPrec_\protoidx) \cdot \prob(\protoidx)
} \label{eq:lgmm_posterior}
\end{equation}
We now turn to the question how to learn a transfer function which maximizes the likelihood of
target space data according to an lGMM in the source space.

\subsection{Linear supervised transfer learning via expectation maximization}

Our second approximation to make the transfer learning problem~\ref{eq:tl} feasible is to
approximate the transfer function $\tartosrc$ with a linear function.
Equivalently, we could say that there exists a matrix $\transf \in \R^{\dimsrc \times \dimtar}$
such that for all $\tardatum_\tardataidx$ in our target space data it holds:
$\tartosrc(\tardatum_\tardataidx) \approx \transf \cdot \tardatum_\tardataidx$.
Another perspective is provided by Saralajew and Villmann, who frame linear supervised
transfer learning as trying to find the first-order Taylor expansion of $\tartosrc$,
while ignoring the constant term \cite{Saralajew2017}.
The linearity restriction may appear rather harsh, but it does have a justification: Transfer learning
is only viable if $\tartosrc$ is \emph{simple} compared to the classifier $\model$.
Otherwise one could simply dismiss the source space model whenever a change in representation
occurs and learn a new model for the target space, as most drift detection approaches do \cite{Ditzler2015}.
Our linearity restriction ensures that $\tartosrc$ has a simple form and
is therefore simple to learn \cite{Paassen2016NC2}. Another advantage of the linearity restriction is
that the function $\tartosrc$ can be parametrized by a single matrix $\transf$.
More precisely, we obtain the following new form for~\ref{eq:tl}, after applying the logarithm:
\begin{equation}
\max_\transf \sum_{\tardataidx = 1}^\tardatalim \log\left[
	\sum_{\protoidx = 1}^\protolim \nDens\big(\transf \cdot \tardatum_\tardataidx \big| \nMean_\protoidx, \nPrec_\protoidx \big)
		\cdot \prob(\lbl_\tardataidx | \protoidx) \cdot \prob(\protoidx)
\right] \label{eq:ml}
\end{equation}

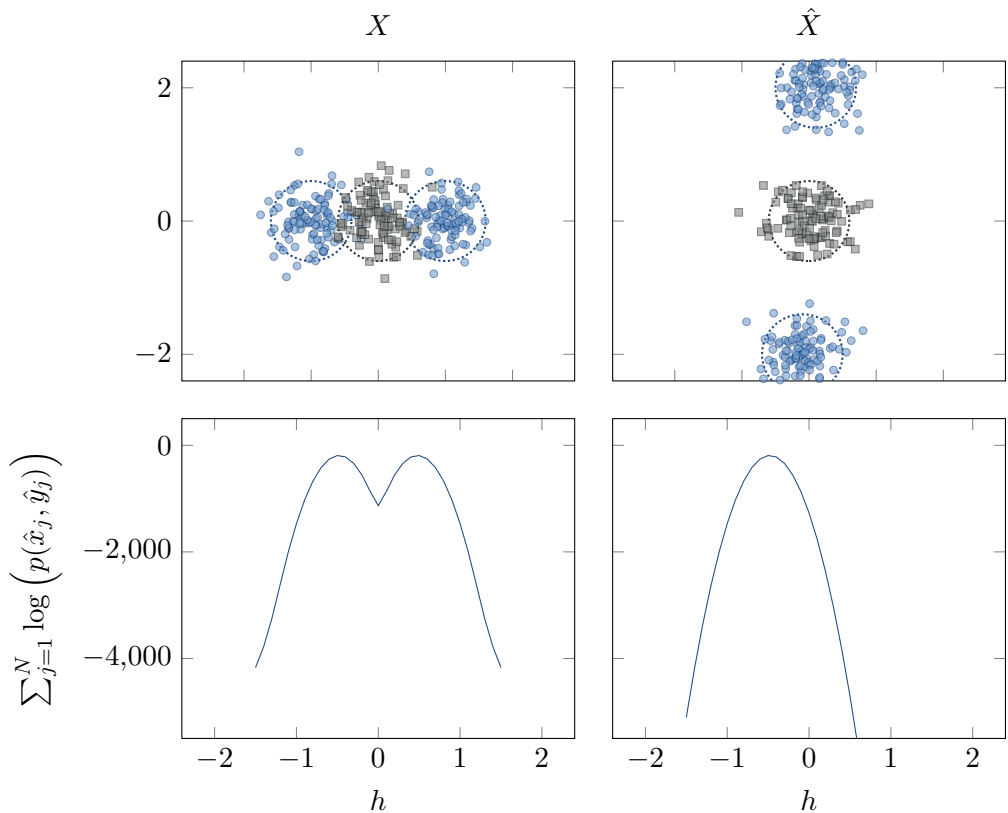
\begin{figure}
\begin{center}
\begin{tikzpicture}
\begin{groupplot}[
	group style={
		group size=2 by 2, horizontal sep=0.5cm, vertical sep=0.5cm,
		x descriptions at=edge bottom,
		y descriptions at=edge left,
	},
	width = 0.45\textwidth,
	scatter/classes={%
		1={mark=*,class1, mark size=0.5mm, opacity=0.6},%
		2={mark=square*,class0, mark size=0.5mm, opacity=0.6},%
		3={mark=*,class1, mark size=0.5mm, opacity=0.6}%
	},%
	enlarge x limits,
	enlarge y limits,
]
\nextgroupplot[disabledatascaling, title = {$\data$}, axis equal, xmin=-2, xmax=2, ymin=-2, ymax=2]
\addplot[scatter,only marks,
scatter src=explicit symbolic]
table[x=x, y=y, meta=label] {toy_data_src.csv};
\draw[densely dotted, class1, fill=none, thick]  (-1, 0) circle (0.6);
\draw[densely dotted, class0, fill=none, thick]  (0, 0) circle (0.6);
\draw[densely dotted, class1, fill=none, thick]  (1, 0) circle (0.6);
\nextgroupplot[disabledatascaling, title = {$\tardata$}, axis equal, xmin=-2, xmax=2, ymin=-2, ymax=2]
\addplot[scatter,only marks,
scatter src=explicit symbolic]
table[x=x, y=y, meta=label] {toy_data_tar.csv};
\begin{scope}[rotate={-90},scale={2}]
\draw[densely dotted, class1, fill=none, thick]  (-1, 0.05) circle (0.3);
\draw[densely dotted, class0, fill=none, thick]  (0, 0) circle (0.3);
\draw[densely dotted, class1, fill=none, thick]  (1, -0.05) circle (0.3);
\end{scope}
\nextgroupplot[xmin=-2, xmax=2, ymin=-5000, ymax=0, xlabel={$h$},
ylabel={$\sum_{\tardataidx=1}^\tardatalim \log\Big(\dens(\tardatum_\tardataidx, \tarlbl_\tardataidx)\Big)$} ]
\addplot[class1color] table[x=h, y=L1] {toy_likelihood.csv};
\nextgroupplot[xmin=-2, xmax=2, ymin=-5000, ymax=0, xlabel={$h$}]
\addplot[class1color] table[x=h, y=L2] {toy_likelihood.csv};
\end{groupplot}
\end{tikzpicture}
\end{center}
\caption{Top left and right: A transfer learning problem with an ambiguous transfer mapping. The
data has been generated as in \fig\ \ref{fig:toy_data} with the difference that the data clusters
to the left and right in the left figure and the top and bottom in the right figure share the same
label.
Bottom left: The log-likelihood according to equation~\ref{eq:ml} of transfer mappings of the form
$\transf = (0, h ; 0, 0)$. $h$ is shown on the x-axis, while the log-likelihood is shown on the
y-axis. As can be seen, the log-likelihood has two local optima.
Bottom right: The log-likelihood if the generating model is constructed as in \fig\ \ref{fig:toy_data}.
In this case, there is only one global optimum.}
\label{fig:toy_likelihood}
\end{figure}

Note that even this restricted problem is challenging. Consider the example in \fig\ 
\ref{fig:toy_likelihood} (top). The data is generated with an lGMM with the same parameters as in \fig\
\ref{fig:toy_data}, except that the label distribution $\prob(\lbl | \protoidx)$ is now given as
$\prob(1 | 1) = \prob(2 | 2) = \prob(1 | 3) = 1$ and $0$ otherwise, implying that the label $3$ is
not generated anymore. In this case, the transfer mapping $\tartosrc$ could either rotate the target
data $\tardata$ to the left \emph{or} to the right, resulting in the same likelihood (see \fig\
\ref{fig:toy_likelihood}, bottom left). In contrast, for the non-ambiguous labeling in
\fig\ \ref{fig:toy_data}, this problem does not occur. In this case, the likelihood has a single
global optimum (see \fig\ \ref{fig:toy_likelihood}, bottom right). This illustrates how label
information contributes crucial disambiguating information for transfer learning.

The problem of finding a (local) optimum for $\transf$ with respect to the log-likelihood in
equation~\ref{eq:ml} can be addressed using an expectation maximization (EM) scheme, as
proposed by Dempster, Laird and Rubin \cite{Dempster1977}. Note that we refer to EM here as a
general optimization scheme for parameters under latent variables, not as a specific optimization
scheme for Gaussian mixture models as discussed by \cite{Bishop2006,Barber2012}. In particular,
we do not intend to adapt the means and covariance matrices of the Gaussian mixture model in our
transfer learning scheme. Instead, we only adapt the transfer matrix $\transf$, which makes the
structure of the problem considerably simpler.

The general EM scheme has two steps, an expectation step and a maximization step.
In the expectation step, we compute the posterior for the latent variables given
the current parameters and in the maximization step we set the parameters in order to maximize the
expected log likelihood $Q$ with respect to the latent variables. The EM scheme starts with
some initial value for the parameters and then iterates the two steps until $Q$ converges.
Dempster, Laird and Rubin have shown that this scheme is guaranteed to achieve a local optimum
for the actual log likelihood \cite{Dempster1977}.

In our case, we intend to optimize the transfer matrix $\transf$, treating the assignment of
data points to Gaussian components as latent variables. Thus, our expectation step is to compute
the posterior for $\protoidx \in \{1, \dots, \protolim\}$ with respect to every transferred test data point
$(\transf \cdot \tardatum_\tardataidx, \tarlbl_\tardataidx)$, while keeping $\transf$ fixed. For compactness,
we denote the posterior $\prob(\protoidx | \transf \cdot \tardatum_\tardataidx, \tarlbl_\tardataidx)$
as $\pstr_{\protoidx | \tardataidx}$. We obtain:
\begin{align}
\pstr_{\protoidx | \tardataidx} &:= 
\prob(\protoidx | \transf \cdot \tardatum_\tardataidx, \tarlbl_\tardataidx) =
\frac{
	\dens(\transf \cdot \tardatum_\tardataidx, \tarlbl_\tardataidx | \protoidx) \cdot \prob(\protoidx)
}{
	\dens(\transf \cdot \tardatum_\tardataidx, \tarlbl_\tardataidx)
} \\
&=
\frac{
	\dens(\transf \cdot \tardatum_\tardataidx | \protoidx) \cdot \prob(\tarlbl_\tardataidx | \protoidx) \cdot \prob(\protoidx)
}{
	\sum_{\protoidx'=1}^\protolim \dens(\transf \cdot \tardatum_\tardataidx | \protoidx') \cdot \prob(\tarlbl_\tardataidx | \protoidx') \cdot \prob(\protoidx')
} \label{eq:tl_posterior}
\end{align}
Note that $\pstr_{\protoidx | \tardataidx}$ can degenerate if a precision matrix $\nPrec_\protoidx$
is not full rank. In such cases, the determinant $\text{det}(\nPrec_\protoidx)$ is $0$
and thus $\nDens(\datum | \nMean_\protoidx, \nPrec_\protoidx) = 0$ is not a valid density.
Such degenerations can be prevented by ensuring that all Eigenvalues of $\nPrec_\protoidx$
stay above some minimum value \cite{Bishop2006,Barber2012} or by replacing the determinant
by the pseudo-determinant, which is defined as the product of all non-zero Eigenvalues.
The former strategy assigns some relevance to all dimensions of the data space and
thus is better able to identify outliers, while the latter strategy entirely disregards
dimensions in which no data variance occurs. Subsequently, we will generally assume that
one of these two strategies is applied to treat degenerate densities.

In the maximization step, we keep the posterior values $\pstr_{\protoidx | \tardataidx}$ fixed and adapt
$\transf$ in order to adapt the expected log likelihood $Q$ with respect to the latent variables.
We denote the adapted transfer matrix as $\transf$ and the transfer matrix from the previous
iteration as $\transf^\text{old}$. Then, we obtain for the maximization step:
\begin{align}
\max_{\transf} Q(\transf | \transf^\text{old}) &= \max_{\transf} \sum_{\tardataidx = 1}^\tardatalim
	\expec_{\protoidx | \transf^\text{old} \cdot \tardatum_\tardataidx, \tarlbl_\tardataidx } \Big[
		\log\big( \dens(\transf \cdot \tardatum_\tardataidx, \tarlbl_\tardataidx, \protoidx) \big)
	\Big] \\
&= \max_{\transf} \sum_{\tardataidx = 1}^\tardatalim \sum_{\protoidx = 1}^\protolim
	\pstr_{\protoidx | \tardataidx} \cdot
	\log\big( \dens(\transf \cdot \tardatum_\tardataidx, \tarlbl_\tardataidx, \protoidx) \big) \notag \\
&= \max_{\transf} \sum_{\tardataidx = 1}^\tardatalim \sum_{\protoidx = 1}^\protolim
	\pstr_{\protoidx | \tardataidx} \cdot \Big(
		\log\big[\dens(\transf \cdot \tardatum_\tardataidx | \protoidx)\big]
		+ \log\big[\prob(\tarlbl_\tardataidx | \protoidx)\big]
		+ \log\big[\prob(\protoidx) \big] \Big) \notag
\end{align}
Note that neither $\prob(\tarlbl_\tardataidx | \protoidx)$ nor $\prob(\protoidx)$ depend on
$\transf$, such that we can disregard them for our optimization problem. If we further
plug in equation~\ref{eq:Gaussian} we obtain:
\begin{align}
& \max_{\transf} \sum_{\tardataidx = 1}^\tardatalim \sum_{\protoidx = 1}^\protolim
	\pstr_{\protoidx | \tardataidx} \cdot \log\left[
	\sqrt{\frac{\det(\nPrec_\protoidx)}{(2 \cdot \pi)^\dimsrc}} \cdot
	\exp\Big(-\frac{1}{2} \cdot
		\transp{(\transf \cdot \tardatum - \nMean_\protoidx)} \cdot \nPrec_\protoidx
		\cdot (\transf \cdot \tardatum - \nMean_\protoidx)
	\Big) \right] \notag \\
= &\min_{\transf} \sum_{\tardataidx = 1}^\tardatalim \sum_{\protoidx = 1}^\protolim
	\pstr_{\protoidx | \tardataidx} \cdot
	\transp{(\transf \cdot \tardatum_\tardataidx - \nMean_\protoidx)} \cdot \nPrec_\protoidx \cdot
		(\transf \cdot \tardatum_\tardataidx - \nMean_\protoidx) =: \qerr(\transf) \label{eq:qerr}
\end{align}
Note that this is just a weighted quadratic error between Gaussian means
$\nMean_\protoidx$ and data points $\transf \cdot \tardatum_\tardataidx$. We denote
this quantity as $\qerr(\transf)$. Interestingly, in this form, the
optimization problem becomes convex and therefore has a guaranteed global solution.

\begin{thm} \label{thm:convexity}
$\qerr$ is a convex function. Further, the gradient of $\qerr$ is given as:
\begin{equation}
\grad{\transf} \qerr(\transf) = 2 \cdot \sum_{\protoidx = 1}^\protolim \nPrec_\protoidx \cdot
	\sum_{\tardataidx = 1}^\tardatalim
		\pstr_{\protoidx | \tardataidx} \cdot
		(\transf \cdot \tardatum_\tardataidx - \nMean_\protoidx) \cdot 
			\transp{\tardatum_\tardataidx} \label{eq:gradient}
\end{equation}
\begin{proof}
To define the gradient of a function with respect to a matrix we consider the derivative
with respect to all matrix entries and put these back into a matrix of the original form.
Via this mechanism, we obtain the gradient
\begin{equation}
\grad{\transf} \transp{(\transf \cdot \tardatum_\tardataidx - \nMean_\protoidx)} \cdot \nPrec_\protoidx \cdot
		(\transf \cdot \tardatum_\tardataidx - \nMean_\protoidx)
= 2 \cdot \nPrec_\protoidx \cdot (\transf \cdot \tardatum_\tardataidx - \nMean_\protoidx) \cdot 
			\transp{\tardatum_\tardataidx} 
\end{equation}
which leads straightforwardly to equation~\ref{eq:gradient}.

For the convexity proof we inspect the Hessian of $\qerr(\transf)$. Here, we define
the Hessian as the matrix of second second derivatives with respect to all entries
of $\transf$. This Hessian is given as \cite{Fackler2005}:
\begin{equation}
\hess{\transf} \qerr(\transf) = 2 \cdot \sum_{\protoidx = 1}^\protolim
	\nPrec_\protoidx \otimes \left(\sum_{\tardataidx = 1}^\tardatalim 
		\pstr_{\protoidx | \tardataidx} \cdot
		\tardatum_\tardataidx \cdot \transp{\tardatum_\tardataidx} \right) \label{eq:Hessian}
\end{equation}
where $\otimes$ is the Kronecker product of two matrices. Recall that $\nPrec_\protoidx$ is
a positive definite matrix and note that $\sum_{\tardataidx = 1}^\tardatalim \pstr_{\protoidx | \tardataidx}
\cdot \tardatum_\tardataidx \cdot \transp{\tardatum_\tardataidx}$ is a positive (semi-)definite matrix.
Further, the Kronecker product of positive (semi-)definite matrices, as well as the sum of positive
(semi-)definite matrices is also guaranteed to be positive (semi-)definite \cite{Fackler2005}.
Therefore, the Hessian of $\qerr(\transf)$ is positive (semi-)definite, which
shows that $\qerr$ is convex.
\end{proof}
\end{thm}

Therefore, we can find a global optimum of $\qerr$ efficiently, using some gradient-based
solver, such as the limited memory Broyden-Fletcher-Goldfarb-Shanno algorithm (l-BFGS).
An even more efficient optimization is possible if all Gaussian components share the same
covariance matrix. In this case, we can provide an analytic solution to~\ref{eq:qerr}.

\begin{thm} \label{thm:closed_form}
Let $\tardata = (\tardatum_1, \ldots, \tardatum_\tardatalim) \in \R^{\dimtar \times \tardatalim}$,
$\pstrmat \in \R^{\protolim \times \tardatalim}$ with
$\pstrmat_{\protoidx \tardataidx} = \pstr_{\protoidx | \tardataidx}$, and
$\protos = (\nMean_1, \ldots, \nMean_\protolim) \in \R^{\dimsrc \times \protolim}$.

If the matrix $\tardata \cdot \transp{\tardata}$ is full rank and there is a matrix
$\nPrec \in \R^{\dimsrc \times \dimsrc}$, such that for all 
$\protoidx \in \{1, \ldots, \protolim\}$ it holds $\nPrec_\protoidx = \nPrec$,
then $\qerr(\transf)$ has a global optimum at
\begin{equation}
\transf = \protos \cdot \pstrmat \cdot \pInv{\tardata} .
\label{eq:closed_form}
\end{equation}
where $\pInv{\tardata} = \transp{\tardata} \cdot (\tardata \cdot \transp{\tardata})^{-1}$ is the
pseudo-Inverse of $\tardata$.

\begin{proof}
As we have shown above, $\qerr(\transf)$ is convex, such that it is sufficient to find
a point $\transf$ with $\grad{\transf} \qerr(\transf) = 0$ for a global optimum.
Such a point can be obtained as follows, starting from equation~\ref{eq:gradient}:
\begin{align}
&&2 \cdot \sum_{\protoidx = 1}^\protolim \nPrec \cdot \sum_{\tardataidx = 1}^\tardatalim
	\pstr_{\protoidx | \tardataidx}
	\cdot (\transf \cdot \tardatum_\tardataidx - \nMean_\protoidx) \cdot 
		\transp{\tardatum_\tardataidx} &= 0 \\
&\iff &\nPrec \cdot \transf \cdot \sum_{\tardataidx = 1}^\tardatalim
	\tardatum_\tardataidx \cdot \transp{\tardatum_\tardataidx}
	&= \nPrec \cdot \sum_{\protoidx = 1}^\protolim 
	 \sum_{\tardataidx = 1}^\tardatalim \pstr_{\protoidx | \tardataidx} \cdot
	\nMean_\protoidx \cdot \transp{\tardatum_\tardataidx} \\
&\iff &\nPrec \cdot \transf \cdot \tardata \cdot \transp{\tardata}
	&= \nPrec \cdot \protos \cdot \pstrmat \cdot \transp{\tardata} \\
&\Leftarrow &\transf &= \protos \cdot \pstrmat \cdot \pInv{\tardata}
\end{align}
\end{proof}
\end{thm}

Note that we can avoid the problem of a rank-deficient matrix $\tardata \cdot \transp{\tardata}$ 
by adding a small positive constant $\regul$ to the diagonal, which ensures full rank and
corresponds to adding the regularization term $\regul \cdot \text{trace}(\nPrec \cdot \transf \cdot \transp{\transf})$
to the error $\qerr(\transf)$ in equation~\ref{eq:qerr}.
Note the striking similarity to the Gaussian prior for linear regression \cite{Bishop2006}.

Further note the $\Leftarrow$ on the last line of the proof. The implication is uni-directional
for this step because the equation $\grad{\transf} \qerr(\transf) = 0$ may have infinitely many solutions
if $\nPrec$ is not full rank. In this case, we could add an arbitrary matrix to $\transf$ which is
constructed from vectors in the null-space of $\nPrec$. Such matrices $\transf'$ would still
be globally optimal solutions. Equation~\ref{eq:closed_form} provides us with just \emph{one} of those
solutions.

The overall expectation maximization algorithm is displayed in algorithm~\ref{alg:em_local}.

\begin{algorithm}
\caption{An expectation maximization algorithm for linear supervised transfer learning on
labeled Gaussian mixture models (lGMMs) with means $\nMean_\protoidx$ and covariance
matrices $\nPrec_\protoidx$ for $\protoidx \in \{1, \ldots, \protolim\}$. The algorithm
gets labeled target space data points $(\tardatum_\tardataidx, \tarlbl_\tardataidx)$
and an error threshold $\epsilon$ as input. The transfer matrix is initialized as
$\eye^{\dimsrc \times \dimtar}$ meaning the $\min\{\dimsrc, \dimtar\}$-dimensional unit matrix,
padded with zeros where necessary.}
\label{alg:em_local}
\begin{algorithmic}
\State $E \gets \infty$, $\transf \gets \eye^{\dimsrc \times \dimtar}$
\While{true}
	\For{$\protoidx \in \{1, \ldots, \protolim\}$}
		\For{$\tardataidx \in \{1, \ldots, \tardatalim\}$}
			\State Compute $\pstr_{\protoidx | \tardataidx}$ according to equation~\ref{eq:tl_posterior}. \Comment Expectation
		\EndFor
	\EndFor
	\If{$\nPrec_\protoidx = \nPrec_{\protoidx'}$ for all $\protoidx, \protoidx'$}
		\State $\transf \gets \protos \cdot \pstrmat \cdot \pInv{\tardata}$ (theorem~\ref{thm:closed_form}). \Comment Maximization
	\Else
		\State Solve problem~\ref{eq:qerr} using a gradient-based solver \Comment Maximization
		\State and gradient~\ref{eq:gradient}. 
	\EndIf
	\State $E' \gets \qerr(\transf)$.
	\If{$|E - E'| < \epsilon$}
		\State \Return $\transf$.
	\EndIf
	\State $E \gets E'$.
\EndWhile
\end{algorithmic}
\end{algorithm}

With regards to computational complexity, we analyze the expectation step and the maximization
step separately. The expectation step requires the computation of $\tardatalim \cdot \protolim$
values of the likelihood $\dens(\transf \cdot \tardatum_\tardataidx | \protoidx)$ according to
equation~\ref{eq:tl_posterior}. Each of these computations is possible in constant time, if we
treat the number of dimensions $\dimtar$ and $\dimsrc$ as constants.
For the maximization step, we need to consider two different cases. If
$\nPrec_\protoidx = \nPrec_{\protoidx'}$ for all $\protoidx, \protoidx'$, we require 
$\effic(\tardatalim \cdot \protolim)$ computations for the matrix product
$\protos \cdot \pstrmat \cdot \pInv{\tardata}$ according to theorem~\ref{thm:closed_form}.
Otherwise, we require an unconstrained optimization algorithm to solve the convex
problem~\ref{eq:qerr}. Such an algorithm will typically need to compute gradients according to
equation~\ref{eq:gradient}, which takes $\effic(\tardatalim \cdot \protolim)$ computations each time.
As the convex optimization problem~\ref{eq:qerr} is just quadratic in nature, we will assume that
a viable optimization algorithm can solve it in a constant number of gradient computations.
Therefore, we obtain $\effic(\tardatalim \cdot \protolim \cdot T)$ for the computational complexity
of algorithm~\ref{alg:em_local}, where $T$ is the number of iterations it takes until the error does
not change more than $\epsilon$ anymore.
It is relatively simple to show that $T$ must be finite (for $\epsilon > 0$) because the
expectation maximization scheme never decreases the likelihood \cite{Barber2012} and the
likelihood is bounded by $1$. In our experimental evaluation, we find that $T$ is typically rather
small (less than 30 iterations) and that the EM approach is therefore considerably faster compared
to the alternatives of learning a new lGMM model or learning a transfer function on the non-convex
GLVQ cost function.
However, providing a theory-grounded estimate for $T$ is challenging. Still, we can gain
some insight by analyzing the special case of a one-to-one assignment of
components to labels, meaning that for each label $\lbl$ there exists exactly one component
$\protoidx$ such that $\prob(\lbl | \protoidx) = 1$. In this case, $\pstr_{\protoidx | \tardataidx}$
is independent of $\transf$ because data points are always assigned crisply to the
component with the matching label. Therefore, the first maximization step directly identifies the
global maximum and in the second iteration the error will not change anymore, yielding $T = 2$
(as is the case in the data from \fig\ \ref{fig:toy_data}). As a rule of thumb, $T$ increases with
the ambiguity in the assignment of data points to components. If there are many components which
can generate a data point, finding a good assignment may take many iterations.
Therefore, it is beneficial in terms of runtime to use as little components as possible, as well as
\enquote{crisp} label distributions, ideally $\prob(\lbl | \protoidx) = 1$ for some label $\lbl$.
The latter point motivates the use of models of the learning vector quantization family, which
feature such crisp assignments.

\subsection{EM Transfer Learning for Learning Vector Quantization Models}

If we already have trained a classifier $\model$ in the source space, training an additional
labeled Gaussian mixture model (lGMM) for transfer learning may appear as unnecessary overhead.
Fortunately, it is possible to obtain a viable lGMM from an existing classifier model.
In particular, an lGMM can be obtained based on a learning vector quantization model.

Learning vector quantization (LVQ) models describe data in terms of \emph{prototypes} $\nMean_\protoidx
\in \srcspace$ which are assigned to a label $\protolbl_\protoidx \in \{1, \ldots, \lbllim\}$.
Data points are classified by assigning the label of the closest prototype, that is:
$\model(\datum) = \argmin_{\lbl} \min_{\protoidx : \protolbl_\protoidx = \lbl} \dist(\datum, \nMean_\protoidx)$
where $\dist$ is the Euclidean distance \cite{Kohonen1995}. A probabilistic
variant of LVQ is \emph{robust soft learning vector quantization} (RSLVQ)
which connects LVQ with Gaussian mixture models \cite{Seo2003}.
RSLVQ essentially formulates a labeled Gaussian mixture model where each component generates
only one label, that is, $\prob(\lbl | \protoidx) = 1$ for one label $\lbl = \protolbl_\protoidx$
and $0$ otherwise.
Also, RSLVQ assumes radial precision matrices 
$\nPrec_\protoidx = \frac{1}{\nDev_\protoidx^2} \eye^\dimsrc$ for some positive scalar
$\nDev_\protoidx$. The positions of the
means $\nMean_\protoidx$ are adjusted according to a stochastic gradient ascent
on the log-posterior $\log(\prob(\lbl | \datum)) = \log[\dens(\datum , \lbl) / \prob(\lbl)]$.
Seo and Obermayer point out that the posterior $\prob(\lbl | \datum)$ for such a model
becomes a crisp winner-takes-all rule if $\nDev_\protoidx$ becomes small for all $\protoidx$
and therefore an RSLVQ becomes a classic LVQ model with prototypes
$\nMean_\protoidx$ \cite{Seo2003}.
Schneider has extended the standard RSLVQ scheme to also permit and learn full precision
matrices $\nPrec_\protoidx$, yielding a full labeled Gaussian mixture model \cite{Schneider2009RSLVQ}.

However, even if the prototypes and precision matrices have not been learned in a
probabilistically motivated fashion, we can use the connection between LVQ models and
lGMM models for the purpose of transfer learning. As an example, consider a local
generalized matrix learning vector quantization (LGMLVQ) model \cite{Schneider2009}.
Such a model is trained via a stochastic gradient descent on the cost function
\begin{equation}
\err = \sum_{\dataidx=1}^\datalim \nonlin\left(
	\frac{
		\dist^2(\nMean^+, \datum_\dataidx) - \dist^2(\nMean^-, \datum_\dataidx)
	}{
		\dist^2(\nMean^+, \datum_\dataidx) - \dist^2(\nMean^-, \datum_\dataidx)
	}
\right) \label{eq:glvq_costfun}
\end{equation}
where $\nMean^+$ is the closest prototype to $\datum_\dataidx$ with the same label,
$\nMean^-$ is the closest prototype to $\datum_\dataidx$ with a different label,
and $\nonlin$ is some sigmoid function.
Importantly, the training does not only adjust the position of the prototypes
$\nMean_\protoidx$ but also adapts the distance $\dist^2(\nMean_\protoidx, \datum_\dataidx) =
\transp{(\nMean_\protoidx - \datum_\dataidx)} \cdot \transp{\relmat_\protoidx}
\cdot \relmat_\protoidx \cdot (\nMean_\protoidx - \datum_\dataidx)$ and learning
the matrix $\relmat_\protoidx$ \cite{Schneider2009}. A notable special case
of LGMLVQ is GMLVQ, in which all prototype share the same matrix $\relmat$ \cite{Schneider2009}.
Also note that LGMLVQ includes generalized learning vector quantization (GLVQ)
as a special case by restricting $\relmat_\protoidx$ to the identity matrix.

Assume now that we have trained a LGMLVQ model in the source space with prototypes
$\nMean_1, \ldots, \nMean_\protolim$ with labels $\protolbl_1, \ldots, \protolbl_\protolim$
and matrices $\relmat_1, \ldots, \relmat_\protolim$.
Then, we obtain an lGMM from this model by setting the Gaussian means to the prototypes,
the precision matrices to $\nPrec_\protoidx = \frac{1}{\nDev^2} \transp{\relmat_\protoidx} \cdot \relmat_\protoidx$,
which is guaranteed to be symmetric and positive semi-definite for any positive scalar $\nDev$.
Further, as in RSLVQ, we set $\prob(\lbl | \protoidx) = 1$ if $\lbl = \protolbl_\protoidx$
and $0$ otherwise \cite{Seo2003}. Finally, we set $\prob(\protoidx) = \frac{1}{\protolim}$
for all $\protoidx$.

As in RSLVQ, we can argue that the resulting lGMM classifies data points in the same
way as the underlying LGMLVQ model and is therefore consistent with it.
This is because the posterior in equation~\ref{eq:lgmm_posterior} collapses to:
\begin{equation}
\prob(\lbl | \datum)
= \frac{
	\sum_{\protoidx : \protolbl_\protoidx = \lbl} \dens(\datum | \protoidx)
}{
	\sum_{\protoidx = 1}^\protolim \dens(\datum | \protoidx)
}
= \frac{
	\sum_{\protoidx : \protolbl_\protoidx = \lbl}
		\sqrt{\det(\nPrec_\protoidx)} \cdot
		\exp\Big(-\frac{1}{2\nDev^2} \cdot \dist^2(\nMean_\protoidx, \datum)\Big)
}{
	\sum_{\protoidx = 1}^\protolim
		\sqrt{\det(\nPrec_\protoidx)} \cdot
		\exp\Big(-\frac{1}{2\nDev^2} \cdot \dist^2(\nMean_\protoidx, \datum)\Big)
}
\end{equation}
Let now $\dist_\protoidx := \dist^2(\nMean_\protoidx, \datum)$ and assume that
$\nMean_{\protoidx^+}$ is the closest prototype to $\datum$, that is, $\dist_{\protoidx^+} < \dist_{\protoidx}$
for all $\protoidx \neq \protoidx^+$.
By multiplying both enumerator and denominator with $\exp(\frac{1}{2\nDev^2} \dist_{\protoidx^+})$ we obtain:
\begin{equation}
\prob(\protolbl_{\protoidx^+} | \datum)
= \frac{\sqrt{\det(\nPrec_{\protoidx^+})} +
	\sum_{\substack{\protoidx \neq \protoidx^+ \\ \protolbl_\protoidx = \protolbl_{\protoidx^+}}}
		\sqrt{\det(\nPrec_\protoidx)} \cdot
		\exp\Big(-\frac{1}{2\nDev^2} \cdot (\dist_\protoidx - \dist_{\protoidx^+})\Big)
}{\sqrt{\det(\nPrec_{\protoidx^+})} +
	\sum_{\protoidx \neq \protoidx^+}
		\sqrt{\det(\nPrec_\protoidx)} \cdot
		\exp\Big(-\frac{1}{2\nDev^2} \cdot (\dist_\protoidx - \dist_{\protoidx^+})\Big)}
\end{equation}
where $\dist_\protoidx - \dist_{\protoidx^+} > 0$ for all $\protoidx \neq \protoidx^+$.
Therefore, $\exp\Big(-\frac{1}{2\nDev^2} \cdot (\dist_\protoidx - \dist_{\protoidx^+})\Big)$
exponentially approaches $0$ for smaller $\nDev$, which leads us to $\prob(\protolbl_{\protoidx^+} | \datum) \approx 1$.
In other words, the maximum a posteriori estimate of the label for $\datum$ is
to assign the label of the closest prototype, which is equivalent to the
classification by LGMLVQ.

\section{Experiments}

In this section, we validate our proposed transfer learning scheme experimentally on three data sets,
two artificial ones and one from the domain of bionic hand prosthesis control. For each data set,
we train a generalized matrix learning vector quantization model (GMLVQ) or a local GMLVQ (LGMLVQ) model
on the source space data. Then, we try to apply the source space model to target space data
via our proposed expectation maximization (EM) transfer learning algorithm. 
We compare the mean classification error obtained by EM with the following reference methods:
\begin{itemize}
\item naively applying the source space model to the target data (\emph{naive}),
\item re-training a new model solely on the target space data (\emph{retrain}),
\item supervised linear transfer learning based on the GLVQ cost function \cite{Paassen2016NC2},
which learns a linear mapping between target and source data using stochastic gradient descent on
the GLVQ cost function~\ref{eq:glvq_costfun} (\emph{GMLVQ}), and
\item utilizing the adaptive support vector machine (\cite{Yang2007}), which tries to apply the
source space classifier to the target space data but corrects wrong predictions using a support
vector machine (\emph{a-SVM}).
\end{itemize}
All implementations are available in our toolbox \cite{EM-TL-Toolbox}.

We analyze the ability for transfer learning of all methods in different conditions: First, we vary
the number of target space training data points available for transfer learning. Our hypothesis is
that our proposed EM transfer learning scheme should require less data to achieve a good
classification error compared to re-training a new model because it only needs to learn a simple
linear transformation, compared to a potentially complex, non-linear classification model (H1).

Second, we remove an entire class from the target space training data. This is particularly
interesting in domains where recording an additional class implies significant additional effort.
We hypothesize that our proposed EM transfer learning scheme should be less affected by missing
classes compared to all other transfer learning approaches (H2) because its focus lies on a change
in representation, not in constructing a model of the class distribution itself. If the
representation change can be estimated using a subset of the classes, any class not contained in
that subset should be omissible without negative effect.

For each experimental condition we report the mean classification test error on the target space
data in a crossvalidation. Finally, we report the runtime of all transfer learning approaches
running on a Intel Core i7-7700 HQ CPU. We expect that our proposed transfer learning approach
will be considerably faster compared to re-training a new model, the a-SVM and GMLVQ transfer
learning (H3), because it involves only a convex optimization for a linear transformation matrix
with fairly few parameters.

For all significance tests we employ a one-sided Wilcoxon signed rank test.

\subsection{Artificial Data I}

\begin{figure}
\begin{center}
\begin{tikzpicture}
\begin{groupplot}[
	group style={
		group size=3 by 2, horizontal sep=0.5cm, vertical sep=0.8cm,
		x descriptions at=edge bottom,
		y descriptions at=edge left,
	},
	width = 0.40\textwidth,
	scatter/classes={%
		1={mark=*,class1, mark size=0.5mm, opacity=0.6},%
		2={mark=square*,class0, mark size=0.5mm, opacity=0.6},%
		3={mark=diamond*,class2, mark size=0.8mm, opacity=0.6},%
		10={mark=*,class1, mark size=1.2mm, line width=0.3mm},%
		20={mark=square*,class0, mark size=1.2mm, line width=0.3mm},%
		30={mark=diamond*,class2, mark size=1.6mm, line width=0.3mm}%
	},%
	enlarge x limits=0.3,
	enlarge y limits,
]
\nextgroupplot[title = {$\data$}, axis equal, xmin=-2, xmax=2, ymin=-2, ymax=2]
\addplot[scatter,only marks,
scatter src=explicit symbolic]
table[x=x, y=y, meta=label] {toy_data_src.csv};
\addplot[scatter,only marks,
scatter src=explicit symbolic]
table[x=x, y=y, meta=label] {toy_protos.csv};
\nextgroupplot[title = {$\tardata$}, axis equal, xmin=-2, xmax=2, ymin=-2, ymax=2]
\addplot[scatter,only marks,
scatter src=explicit symbolic]
table[x=x, y=y, meta=label] {toy_data_tar.csv};
\addplot[scatter,only marks,
scatter src=explicit symbolic]
table[x=x, y=y, meta=label] {toy_protos.csv};
\nextgroupplot[title = {$\transf \cdot \tardata$}, axis equal, xmin=-2, xmax=2, ymin=-2, ymax=2]
\addplot[scatter,only marks,
scatter src=explicit symbolic]
table[x=x, y=y, meta=label] {toy_data_transfer.csv};
\addplot[scatter,only marks,
scatter src=explicit symbolic]
table[x=x, y=y, meta=label] {toy_protos.csv};
\nextgroupplot[title = {$\relmat \cdot \data$}, height=2cm, %
	xmin=-2, xmax=2, ymin=-0.5, ymax=0.5, hide y axis]
\addplot[scatter,only marks,
scatter src=explicit symbolic]
table[x=x, y=y, meta=label] {toy_data_src_omega.csv};
\addplot[scatter,only marks,
scatter src=explicit symbolic]
table[x=x, y=y, meta=label] {toy_protos_omega.csv};
\nextgroupplot[title = {$\relmat \cdot \tardata$}, height=2cm,%
	xmin=-2, xmax=2, ymin=-0.5, ymax=0.5, hide y axis]
\addplot[scatter,only marks,
scatter src=explicit symbolic]
table[x=x, y=y, meta=label] {toy_data_tar_omega.csv};
\addplot[scatter,only marks,
scatter src=explicit symbolic]
table[x=x, y=y, meta=label] {toy_protos_omega.csv};
\nextgroupplot[title = {$\relmat \cdot \transf \cdot \tardata$}, height=2cm,%
	xmin=-2, xmax=2, ymin=-0.5, ymax=0.5, hide y axis]
\addplot[scatter,only marks,
scatter src=explicit symbolic]
table[x=x, y=y, meta=label] {toy_data_transfer_omega.csv};
\addplot[scatter,only marks,
scatter src=explicit symbolic]
table[x=x, y=y, meta=label] {toy_protos_omega.csv};
\end{groupplot}
\end{tikzpicture}
\end{center}
\caption{A visualization of a GMLVQ model trained on the toy data set shown in \fig\ \ref{fig:toy_data}.
Prototypes are highlighted via bigger size. Shape and color indicate the class assignment.
Top row: The source space data $\data$, target space data $\tardata$ and transferred target space
data $\transf \cdot \tardata$.
Bottom row: The same data after multiplication via the GMLVQ relevance matrix $\relmat$.}
\label{fig:toy_data_model}
\end{figure}
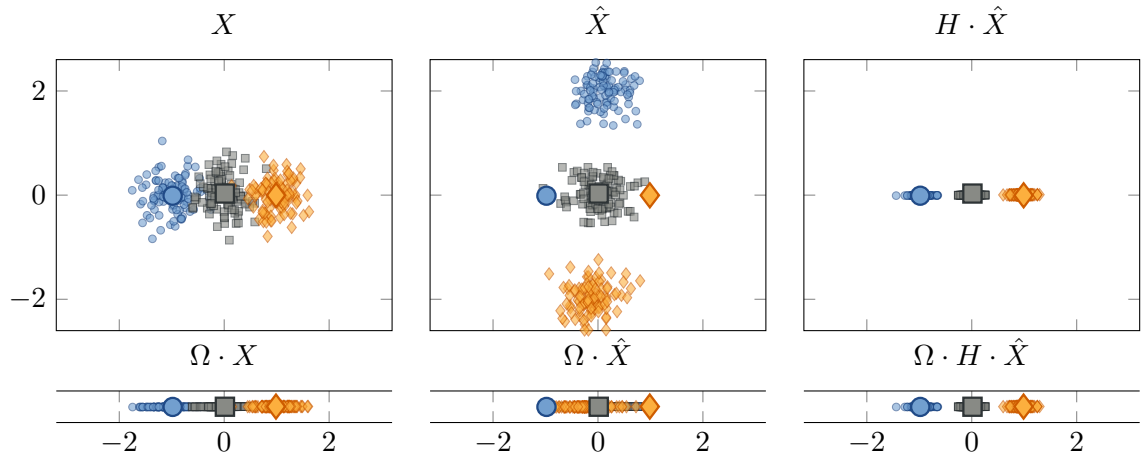

Our first data set is the two-dimensional toy data set shown in \fig\ \ref{fig:toy_data}.
The data is generated via a labeled Gaussian mixture model
with one component for each of the three classes with means $\nMean_1 = (-1, 0)$,
$\nMean_2 = (0, 0)$ and $\nMean_3 = (1, 0)$ and shared covariance matrix
$\nPrec^{-1} = 0.3^2 \cdot \eye^2$.

The target data is generated with a similar model but with the
means set to $\nMean_1 = (-0.1, -2)$, $\nMean_2 = (0, 0)$, and $ \nMean_3 = (0.1, 2)$.
Such a rotation of the data is similar to
the effect of electrode shifts in the domain of bionic hand prostheses \cite{Khushaba2014}.
In both source and target space we generate $100$ data points per class.

As a source space model, we employ a GMLVQ
model with one prototype per class, which is shown in \fig\ \ref{fig:toy_data_model} (top left).
On the source space data, the GMLVQ model correctly identifies the first dimension as discriminative
and discards the second dimension via the relevance matrix $\relmat$
(see \fig\ \ref{fig:toy_data_model}, bottom left). However, for the target space data this model
is invalid because the second dimension now carries the discriminative information 
(see \fig\ \ref{fig:toy_data_model}, middle column). Quantitatively, we obtain a classification
error on the target space data of above $60\%$ (see \fig\ \ref{fig:toy_data_results}, left).

\pgfplotsset{
	src/.style={%
		fill=scarletred1,
		draw=scarletred3, thick,
		ybar,
		error bars/.cd,
		y dir=both,y explicit,
		error bar style={scarletred3, thick}%
	},
	tar/.style={%
		fill=skyblue1,
		draw=skyblue3, thick,
		ybar,
		error bars/.cd,
		y dir=both,y explicit,
		error bar style={skyblue3, thick}
	},
	errline/.style={%
		error bars/.cd,
		y dir=both,y explicit
	},
	em/.style={chameleon3, mark=o, errline},
	em_loc/.style={chameleon2, mark=o, errline},
	retrain/.style={plum3, xshift=1pt, mark=square, errline},
	retrain_loc/.style={plum2, xshift=1pt, mark=square, errline},
	gmlvq/.style={orange3, xshift=2pt, mark=triangle, mark size=0.9mm, errline},
	asvm/.style={chocolate3, xshift=3pt, mark=diamond, mark size=0.9mm, errline},
	barplot/.style={%
		xmin=0,
		xmax=1,
		xtick={0,1},
		xticklabels={source, naive},
		xticklabel style={rotate=90, anchor=east},
		xlabel={},
		ylabel={avg. error},
		width=2.5cm,
		bar width=0.2cm,
		enlarge x limits=1%
	},
        barplot_loc/.style={%
		xmin=0,
		xmax=1,
		xtick={0,1},
		xticklabels={source-loc, naive-loc},
		xticklabel style={rotate=90, anchor=east},
		xlabel={},
		ylabel={avg. error},
		width=2.5cm,
		bar width=0.2cm,
		enlarge x limits=1%
	},
        barplot_four/.style={%
		xmin=0,
		xmax=3,
		xtick={0,1,2,3},
		xticklabels={source, source-loc, naive, naive-loc},
		xticklabel style={rotate=90, anchor=east},
		xlabel={},
		ylabel={avg. error},
		width=3.25cm,
		bar width=0.2cm,
		enlarge x limits=0.4%
	}
}

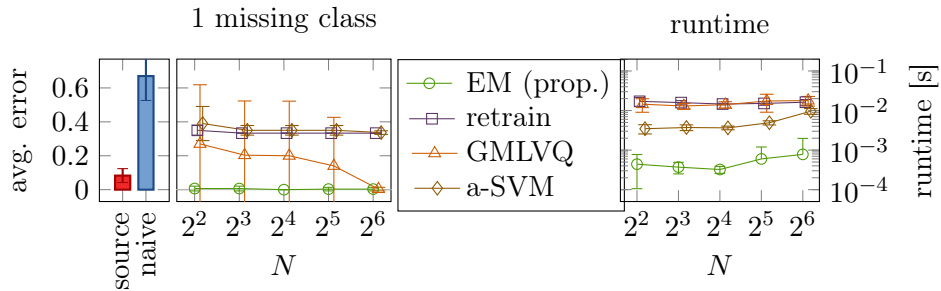
\begin{figure}
\begin{center}
\begin{tikzpicture}
\begin{groupplot}[
	group style={
		group size=4 by 1,
		horizontal sep=0.1cm,
		vertical sep=1.3cm,
		x descriptions at=edge bottom,
	},
	height=3.47cm,
	width=4.4cm,
	xlabel={$\tardatalim$},
	xmode=log,
	log basis x={2},
	xmin=4,
	xmax=64,
	xtick={4,8,16,32,64},
	log basis ticks=2,
	enlarge x limits=0.1,
	enlarge y limits=true,
	ymin=0,
	ymax=0.7,
	ytick={0,0.2,0.4, 0.6},
	legend cell align=left,
	legend pos=outer north east,
]
\nextgroupplot[xmode=linear, barplot]
\addplot[src] plot coordinates {(0,0.083) +- (0,0.041)};
\addplot[tar] plot coordinates {(1,0.67) +- (0,0.144)};
\nextgroupplot[
	title={1 missing class},
	y tick label style={opacity=0},
]
\addplot[aluminium3, sharp plot, ultra thin, update limits=false, forget plot]
coordinates {(0.001,0) (1000,0)};
\addplot[em] table[x=n, y=err_mean_em, y error=err_std_em] {toy_data_errors.csv};
\addlegendentry{EM (prop.)}
\addplot[retrain] table[x=n, y=err_mean_retrain, y error=err_std_retrain] {toy_data_errors.csv};
\addlegendentry{retrain}
\addplot[gmlvq] table[x=n, y=err_mean_gmlvq, y error=err_std_gmlvq] {toy_data_errors.csv};
\addlegendentry{GMLVQ}
\addplot[asvm] table[x=n, y=err_mean_a-svm, y error=err_std_a-svm] {toy_data_errors.csv};
\addlegendentry{a-SVM}
\nextgroupplot[group/empty plot]
\nextgroupplot[
	title={runtime},
	ylabel={runtime [s]},
	ymode=log,
	ymin=0.0001,
	ymax=0.1,
	ytick={0.0001, 0.001, 0.01,0.1},
	yticklabel pos=right,
	width=4.2cm,
]
\addplot[em] table[x=n, y=time_mean_em, y error=time_std_em] {toy_data_runtimes.csv};
\addplot[retrain] table[x=n, y=time_mean_retrain, y error=time_std_retrain] {toy_data_runtimes.csv};
\addplot[gmlvq] table[x=n, y=time_mean_gmlvq, y error=time_std_gmlvq] {toy_data_runtimes.csv};
\addplot[asvm] table[x=n, y=time_mean_a-svm, y error=time_std_a-svm] {toy_data_runtimes.csv};
\end{groupplot}
\end{tikzpicture}
\end{center}
\caption{Mean classification error (left, middle) and mean runtime (right) in a
ten-fold crossvalidation on the toy data set shown in \fig\ \ref{fig:toy_data}
with the left and middle class being available for transfer learning. The $x$-axis indicates the
number of available target space training data points $\tardatalim$ (in log scaling) while the
$y$-axis displays the mean classification error (left, middle) or the runtime (right, log scale).
Error bars indicate the standard deviation.}
\label{fig:toy_data_results}
\end{figure}

For this data set, we only consider the case in which the right class 
(diamonds in \fig\ \ref{fig:toy_data_model}) is \emph{not} available in the target space training
data. The mean classification error and runtime for this case are shown in
\fig\ \ref{fig:toy_data_results}. Even if only four data points are available as training data,
the proposed EM transfer learning algorithm consistently identifies a viable transfer mapping
$\transf$ such that the source model achieves a classification error below $1\%$ (middle).
To achieve the same consistency, GMLVQ transfer learning requires at least $2^6 = 64$ data points,
while both a-SVM and retrain necessarily fail to classify the missing third class, yielding a
classification error of above $30\%$. These results lend support for H1 and H2.

Regarding runtime, we observe that our proposed transfer learning scheme is roughly 10 times faster
compared to a-SVM and roughly 30 times faster compared to GMLVQ transfer learning and learning a
new GMLVQ model (see \fig\ \ref{fig:toy_data_results}, right), supporting H3.

\subsection{Artificial Data II}

\begin{figure}
\begin{center}
\begin{tikzpicture}
\begin{groupplot}[
        disabledatascaling,
	group style={
		group size=2 by 1, horizontal sep=0.5cm, vertical sep=0.8cm,
		x descriptions at=edge bottom,
		y descriptions at=edge left,
	},
	width = 0.525\textwidth,
	scatter/classes={%
		1={mark=*,class0, mark size=0.5mm, opacity=0.6},%
		2={mark=square*,class1, mark size=0.5mm, opacity=0.6},%
		3={mark=diamond*,class2, mark size=0.8mm, opacity=0.6}
	},
	enlarge x limits=0.3,
	enlarge y limits,
]

\nextgroupplot[title = {$\data$}, axis equal, xmin=-2, xmax=2, ymin=-2, ymax=2]
\addplot[scatter,only marks,
scatter src=explicit symbolic]
table[x=x, y=y, meta=label] {cigars_src.csv};

\nextgroupplot[title = {$\tardata$}, axis equal, xmin=-2, xmax=2, ymin=-2, ymax=2]
\addplot[scatter,only marks,
scatter src=explicit symbolic]
table[x=x, y=y, meta=label] {cigars_tar.csv};

\end{groupplot}
\end{tikzpicture}
\end{center}
\caption{The 3 cigars data set consisting of three classes generated from non-spherical Gaussians. 
Source data are shown in the left plot and target data in the right.}
\label{fig:cigars_data}
\end{figure}
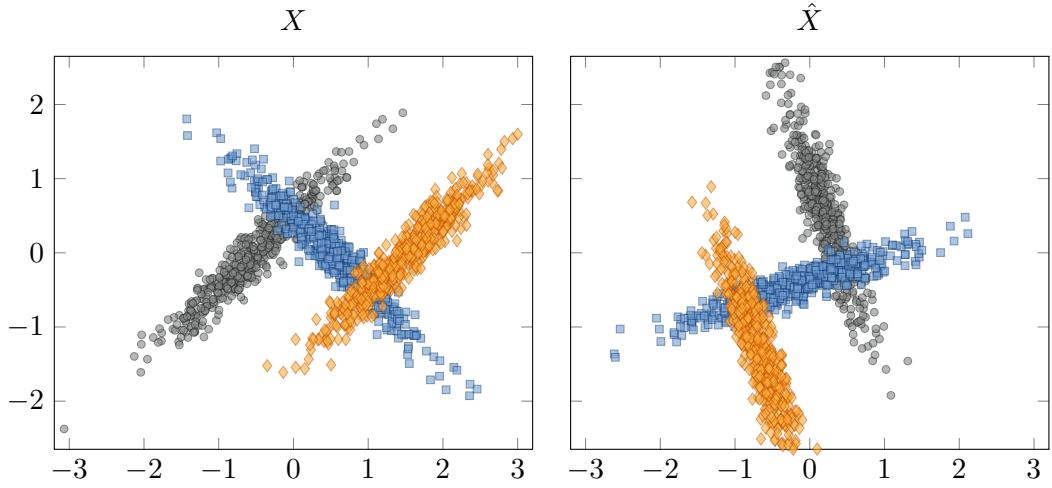

Our second artificial data set illustrates the advantage of individual precision matrices in cases
where strong class overlap is present. The data set is inspired by the \emph{cigars} data set by
\cite{Schneider2009} and consists of $1000$ data points for each of the three classes. The data
is generated via a labeled Gaussian mixture model (lGMM) with one component per class, with means at
$\nMean_1 = (-0.5,0)$, $\nMean_2 = (0.5, 0)$, and $\nMean_3 = (1.5,0)$, and
covariance matrices
\begin{equation*}
\nPrec_1^{-1} = \nPrec_3^{-1} = \begin{pmatrix} 0.485 & 0.36 \\ 0.36 & 0.485 \end{pmatrix}, \qquad
\text{and} \quad \nPrec_2^{-1} = \begin{pmatrix} 0.485 & -0.36 \\ -0.36 & 0.485 \end{pmatrix}
\end{equation*}
The target data is generated from the same distribution,
with the model being rotated by $90^\circ$ (see \fig\ \ref{fig:cigars_data}).

\begin{figure}
\begin{center}
\begin{tikzpicture}
\begin{groupplot}[
	group style={
		group size=4 by 1,
		horizontal sep=0.1cm,
		vertical sep=1.3cm,
		x descriptions at=edge bottom,
	},
	height=3.47cm,
	width=4.4cm,
	xlabel={$\tardatalim$},
	xmode=log,
	log basis x={2},
	xmin=4,
	xmax=64,
	xtick={4,8,16,32,64},
	log basis ticks=2,
	enlarge x limits=0.1,
	enlarge y limits=true,
	ymin=0,
	ymax=0.7,
	ytick={0,0.2,0.4, 0.6},
	legend cell align=left,
	legend pos=outer north east,
        legend style={font=\tiny\selectfont},
]
\nextgroupplot[xmode=linear, barplot_four]
\addplot[densely dashed, src] plot coordinates {(0,0.213) +- (0,0.037)};
\addplot[src] plot coordinates {(1,0.097) +- (0,0.040)};
\addplot[densely dashed, tar] plot coordinates {(2,0.519) +- (0,0.060)};
\addplot[tar] plot coordinates {(3,0.639) +- (0,0.067)};
\nextgroupplot[
        title={1 missing class},
	y tick label style={opacity=0},
]
\addplot[aluminium3, sharp plot, ultra thin, update limits=false, forget plot]
coordinates {(0.001,0) (1000,0)};
\addplot[densely dashed, em] table[x=n, y=err_mean_em, y error=err_std_em] {cigars_data_2_labels_errors.csv};
\addlegendentry{EM (prop.)}
\addplot[em_loc] table[x=n, y=err_mean_em-loc, y error=err_std_em-loc] {cigars_data_2_labels_errors.csv};
\addlegendentry{EM-loc (prop.)}
\addplot[densely dashed, retrain] table[x=n, y=err_mean_retrain, y error=err_std_retrain] {cigars_data_2_labels_errors.csv};
\addlegendentry{retrain}
\addplot[retrain_loc] table[x=n, y=err_mean_retrain-loc, y error=err_std_retrain-loc] {cigars_data_2_labels_errors.csv};
\addlegendentry{retrain-loc}
\addplot[densely dashed, gmlvq] table[x=n, y=err_mean_gmlvq, y error=err_std_gmlvq] {cigars_data_2_labels_errors.csv};
\addlegendentry{GMLVQ}
\addplot[densely dashed, asvm] table[x=n, y=err_mean_a-svm, y error=err_std_a-svm] {cigars_data_2_labels_errors.csv};
\addlegendentry{a-SVM}
\nextgroupplot[group/empty plot]
\nextgroupplot[
        title={runtime},
	ylabel={runtime [s]},
	ymode=log,
	ymin=0.0005,
	ymax=0.5,
	ytick={0.001, 0.01,0.1,1},
	yticklabel pos=right,
	width=4.2cm,
]
\addplot[densely dashed, em] table[x=n, y=time_mean_em, y error=time_std_em] {cigars_data_3_labels_runtimes.csv};
\addplot[em_loc] table[x=n, y=time_mean_em-loc, y error=time_std_em-loc] {cigars_data_3_labels_runtimes.csv};
\addplot[densely dashed, retrain] table[x=n, y=time_mean_retrain, y error=time_std_retrain] {cigars_data_3_labels_runtimes.csv};
\addplot[retrain_loc] table[x=n, y=time_mean_retrain-loc, y error=time_std_retrain-loc] {cigars_data_3_labels_runtimes.csv};
\addplot[densely dashed, gmlvq] table[x=n, y=time_mean_gmlvq, y error=time_std_gmlvq] {cigars_data_3_labels_runtimes.csv};
\addplot[densely dashed, asvm] table[x=n, y=time_mean_a-svm, y error=time_std_a-svm] {cigars_data_3_labels_runtimes.csv};
\end{groupplot}
\end{tikzpicture}
\end{center}
\caption{Mean classification error (left, middle) and mean runtime (right) in a
$30$-fold crossvalidation on the cigars data set shown in \fig\ \ref{fig:cigars_data}
with the left and middle class being available
for transfer learning. The $x$-axis indicates the number of available target
space training data points $\tardatalim$ (in log scaling) while the $y$-axis displays the mean
classification (left middle) or the runtime (right, log scale). Error bars indicate the standard
deviation. Approaches with individual precision matrices per class are marked with \enquote{loc.}.
Approaches with shared precision matrices are drawn with dashed lines.}
\label{fig:cigars_results}
\end{figure}
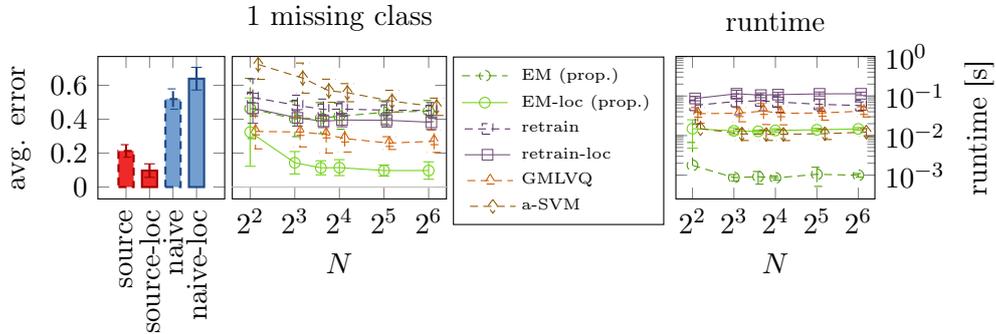

As a source space model, we employ GMLVQ and LGMLVQ with one prototype per class.
The challenge in classifying this data lies in the fact that the discriminative direction,
that is, the direction orthogonal to the main axis of the classes covariance matrix, is different for
the middle class compared to the other two. A lGMM with just one component per class can only account
for this difference by using local precision matrices. Accordingly, we observe that the source classification
error for GMLVQ (i.e.\ a lGMM with shared precision matrix) is much higher compared to LGMLVQ (i.e.\ a lGMM
with local precision matrices per component) with $21.33\%$ versus $9.73\%$ on average (see source and source-loc
in \fig\ \ref{fig:cigars_results}, left).

For transfer learning on this data, we utilize only data of the left and middle class.
The mean classification error across $30$ crossvalidation trials is depicted in
\fig\ \ref{fig:cigars_results} (middle). As expected, we observe that our proposed transfer
learning scheme based on the LGMLVQ model (EM-loc) outperforms all other approaches significantly
if at least 12 data points are available ($p < 10^{-3}$) and achieves an error below $12\%$
on average, close to the error of the source space model. These results lend support for both
H1 and H2.

Regarding runtime we observe that the proposed expectation maximization scheme for individual
precision matrices (EM-loc) is about 10 times slower compared to the scheme for a shared precision
matrix (see \fig\ \ref{fig:cigars_results}, right). This is due to the fact that the latter
approach can exploit a closed-form solution for $\transf$ while the former needs to employ an
iterative solver. This also makes EM-loc about as slow as a-SVM, but still about 3 times faster
compared to GMLVQ transfer learning and about 8 times faster compared to re-training a new
LGMLVQ model on the target data. Thus, H3 is partially supported.

\subsection{Myoelectric data}

The motivation for our myoelectric data set is to learn a classifier mapping from electromyographic
(EMG) recordings to the corresponding hand motion. Such a mapping can be utilized as an user
interface for a hand prosthesis because, after amputation, the hand remains represented in the brain.
By actuating the so called phantom hand, the residual muscles in the stump are activated via the
motor neurons. This leads to corresponding EMG signals that can be mapped to the desired motion,
which is then executed by the prosthesis \cite{Farina2014}.

This data set consists of electromyographic (EMG) recordings of hand motions of $10$ able-bodied
participants, recorded with a high-density grid of $96$ EMG electrodes with $8$ mm inter-electrode
distance, located around the forearm at $1/3$ of the distance from elbow to wrist.

Each participant performed $15$ to $35$ runs ($236$ runs in total) of a series of six hand
movements, namely wrist pronation/supination, wrist flexion/extension and finger spread/fist,
intermitted by resting phases.
Each motion lasted $3$ seconds from which the first and the last second were cut to avoid label
noise, leaving $1$ seconds of each motion for analysis. The experiments are in accordance with the
declaration of Helsinki and approved by the local ethics commission. Further details on the
experimental protocol are provided in \cite{Hahne2012}.

Our classification task is to identify the correct motion corresponding to the current EMG signal,
including an additional resting class (i.e.\ $7$ classes in total).
All signals were filtered with a low pass ($500$ Hz, fourth-order Butterworth), a high pass
($20$ Hz, fourth-order Butterworth), and a band stop filter ($45-55$ Hz, second-order Butterworth)
to remove noise, movement artifacts, and power line interferences respectively.
As features, we employ the logarithm of the signal variance for each electrode, computed on
non-overlapping time windows of $100ms$ length.
Thus, depending on the number of runs, $1925$ to $3255$ samples were available per participant,
balanced for all classes (for the participant with the fewest runs we obtained $275$ samples per
class, for the participant with the most runs $465$ samples per class).

Since high-density EMG recordings are not common in prosthetic hardware \cite{Farina2014}, we
simulate a more realistic setup by using a subset of $8$ equidistant electrodes located on a ring
around the forearm (see figure~\ref{fig:electrode_shift}, top left).
In order to obtain disturbed target data, we simulate an electrode shift by utilizing
eight different electrodes, located one step within the array ($8mm$) transversely to the forearm
(see figure~\ref{fig:electrode_shift}, bottom left).
Such electrode shifts pose a serious problem in real-life prosthesis control,
since they occur frequently, e.g.\ after reapplying the prosthesis, and lead to significantly
decreased classification accuracy \cite{Farina2014}.

As a first analysis, we evaluate which classification method performs best on the source data set.
In particular, we compare a generalized matrix learning vector quantization (GMLVQ),
local GMLVQ (LGMLVQ), a labeled Gaussian mixture model with shared precision matrix (slGMM),
a labeled Gaussian mixture model with individual precision matrices (lGMM), a slGMM with GMLVQ
initialization (GMLVQ + slGMM) and a lGMM with LGMLVQ initialization (LGMLVQ + lGMM). The Gaussian
mixture models were trained with expectation maximization while restricting the standard deviation
in each dimension to be at least $0.001$, as described by \cite{Barber2012}.
For each of the methods, we vary the number of prototypes/Gaussian components from $1$ to $5$.
In our analysis, we iterate over all $236$ runs in the data set and treat the data of the current run
as test data, yielding a leave-one-out crossvalidation over the $236$ runs. As training data we
utilize a random sample of $175$ data points, balanced over the classes, drawn
from the remaining runs of the same subject. We train each model starting from $5$ random
initializations and select the model with the lowest training error. For this model, we then record
the classification error on the test data.

\begin{table}
\begin{center}
\footnotesize
\begin{tabular}{lCCCCCW}
$\protolim$ & GMLVQ    & LGMLVQ            & slGMM                 & lGMM             & GMLVQ + slGMM         & LGMLVQ + lGMM \\
\cmidrule(r){1-1} \cmidrule(r){2-3} \cmidrule(r){4-5} \cmidrule{6-7}
$1$ & $6.7 \pm 7.1 \%$ & $7.0 \pm 7.2 \%$  & $\bm{5.9 \pm 6.7 \%}$ & $6.7 \pm 7.1 \%$ & $\bm{5.9 \pm 6.7 \%}$ & $6.7 \pm 7.1 \%$ \\
$2$ & $6.5 \pm 6.8 \%$ & $8.4 \pm 7.9 \%$  & $5.8 \pm 6.5 \%$      & $6.5 \pm 6.6 \%$ & $\bm{5.6 \pm 6.2 \%}$ & $9.9 \pm 8.0 \%$ \\
$3$ & $6.7 \pm 7.3 \%$ & $9.3 \pm 8.5 \%$  & $6.1 \pm 6.7 \%$      & $7.1 \pm 7.7 \%$ & $\bm{5.7 \pm 6.4 \%}$ & $9.6 \pm 8.7 \%$ \\
$4$ & $6.5 \pm 7.4 \%$ & $9.9 \pm 8.9 \%$  & $\bm{5.9 \pm 6.6 \%}$ & $7.8 \pm 7.4 \%$ & $\bm{5.9 \pm 6.7 \%}$ & $11.9 \pm 12.8 \%$ \\
$5$ & $6.4 \pm 7.4 \%$ & $10.1 \pm 8.8 \%$ & $\bm{5.9 \pm 6.7 \%}$ & $7.8 \pm 7.3 \%$ & $\bm{5.9 \pm 6.4 \%}$ & $23.1 \pm 28.9 \%$ \\
\end{tabular}
\end{center}
\caption{Mean classification test error and standard deviation on the source space data
across all runs in the myoelectric data set. The different classification models are listed on the x axis,
the number of prototypes / Gaussian components for the model on the y axis. The best results in
each row are highlighted via bold print.}
\label{tab:myo_prototypes}
\end{table}

The results of our pre-experiment are shown in \tab~\ref{tab:myo_prototypes}. As can be seen, a
labeled Gaussian mixture model with shared precision matrix (slGMM) and GMLVQ initialization
consistently achieves the best results.
The difference in error is significant compared to GMLVQ ($p < 0.05$), LGMLVQ ($p < 0.001$),
lGMM ($p < 0.001$), and lGMM with LGMLVQ initialization ($p < 0.001$). The difference to a slGMM
without GMLVQ initialization is insignificant. Regarding the number of prototypes we obtain the best
results for $\protolim = 2$ prototypes, although the error difference to other values for
$\protolim$ is insignificant. For the main analysis, we select the overall best model, namely
slGMM with GMLVQ initialization and $\protolim = 2$.

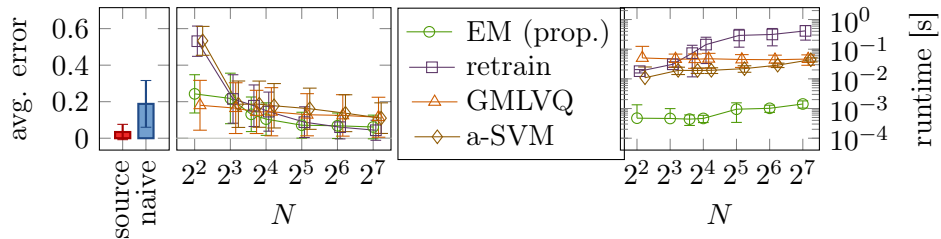
\begin{figure}
\begin{center}
\begin{tikzpicture}
\begin{groupplot}[
	group style={
		group size=4 by 1,
		horizontal sep=0.1cm,
		vertical sep=1.3cm,
		x descriptions at=edge bottom,
	},
	height=3.47cm,
	width=4.4cm,
	xlabel={$\tardatalim$},
	xmode=log,
	log basis x={2},
	xmin=4,
	xmax=128,
	xtick={4,8,16,32,64,128},
	log basis ticks=2,
	enlarge x limits=0.1,
	enlarge y limits=true,
	ymin=0,
	ymax=0.65,
	ytick={0,0.2,0.4, 0.6},
	legend cell align=left,
	legend pos=outer north east,
]
\nextgroupplot[xmode=linear, barplot]
\addplot[src] plot coordinates {(0,0.034) +- (0,0.042)};
\addplot[tar] plot coordinates {(1,0.188) +- (0,0.128)};
\nextgroupplot[
	y tick label style={opacity=0},
]
\addplot[aluminium3, sharp plot, ultra thin, update limits=false, forget plot]
coordinates {(0.001,0) (1000,0)};
\addplot[em] table[x=n, y=err_mean_vq, y error=err_std_vq] {myo_data_errors.csv};
\addlegendentry{EM (prop.)}
\addplot[retrain] table[x=n, y=err_mean_retrain, y error=err_std_retrain] {myo_data_errors.csv};
\addlegendentry{retrain}
\addplot[gmlvq] table[x=n, y=err_mean_gmlvq, y error=err_std_gmlvq] {myo_data_errors.csv};
\addlegendentry{GMLVQ}
\addplot[asvm] table[x=n, y=err_mean_asvm, y error=err_std_asvm] {myo_data_errors.csv};
\addlegendentry{a-SVM}
\nextgroupplot[group/empty plot]
\nextgroupplot[
	ylabel={runtime [s]},
	ymode=log,
	ymin=0.0001,
	ymax=1,
	ytick={0.0001, 0.001, 0.01,0.1,1},
	yticklabel pos=right,
	width=4.2cm,
]
\addplot[em] table[x=n, y=time_mean_vq, y error=time_std_vq] {myo_data_runtimes.csv};
\addplot[retrain] table[x=n, y=time_mean_retrain, y error=time_std_retrain] {myo_data_runtimes.csv};
\addplot[gmlvq] table[x=n, y=time_mean_gmlvq, y error=time_std_gmlvq] {myo_data_runtimes.csv};
\addplot[asvm] table[x=n, y=time_mean_asvm, y error=time_std_asvm] {myo_data_runtimes.csv};
\end{groupplot}
\end{tikzpicture}
\end{center}
\caption{Mean classification error (left, middle) and mean runtime (right) across all runs
in the myoelectric data set.
The $x$-axis indicates the number of available target space training data points $\tardatalim$ (in log scaling)
while the $y$-axis displays the mean classification error (left, middle) or the runtime
(right, log scale). Error bars indicate the standard deviation.}
\label{fig:myo_data_results}
\end{figure}

\begin{table}
\begin{center}
\footnotesize
\begin{tabular}{lWWWWW}
$\tardatalim$ & naive      & EM (prop.)             & retrain               & GMLVQ                   & a-SVM \\
\cmidrule(r){1-1} \cmidrule(r){2-2} \cmidrule(r){3-3} \cmidrule(r){4-4} \cmidrule(r){5-5} \cmidrule{6-6}
$4$   & $18.8 \pm 12.8 \%$ & $24.3 \pm 10.5 \%$     & $53.1 \pm 8.3 \%$     & $\bm{18.0 \pm 13.6 \%}$ & $53.3 \pm 7.9 \%$ \\
$8$   & $18.8 \pm 12.8 \%$ & $21.8 \pm 13.8 \%$     & $21.5 \pm 13.3 \%$    & $\bm{16.5 \pm 13.9 \%}$ & $18.6 \pm 12.7 \%$ \\
$12$  & $18.8 \pm 12.8 \%$ & $\bm{13.1 \pm 9.5 \%}$ & $17.8 \pm 11.4 \%$    & $14.4 \pm 11.8 \%$      & $18.6 \pm 12.7 \%$ \\
$16$  & $18.8 \pm 12.8 \%$ & $\bm{10.5 \pm 8.9 \%}$ & $14.6 \pm 10.7 \%$    & $14.5 \pm 13.2 \%$      & $17.8 \pm 12.2 \%$ \\
$32$  & $18.8 \pm 12.8 \%$ & $\bm{7.1 \pm 7.1 \%}$  & $8.8 \pm 8.4 \%$      & $12.9 \pm 11.5 \%$      & $16.1 \pm 11.3 \%$ \\
$64$  & $18.8 \pm 12.8 \%$ & $6.8 \pm 7.0 \%$       & $\bm{6.0 \pm 6.4 \%}$ & $12.5 \pm 11.3 \%$      & $13.7 \pm 10.1 \%$ \\
$128$ & $18.8 \pm 12.8 \%$ & $6.2 \pm 6.5 \%$       & $\bm{4.4 \pm 5.5 \%}$ & $11.5 \pm 10.9 \%$      & $11.0 \pm 8.4 \%$ \\
\end{tabular}
\end{center}
\caption{Mean classification test error and standard deviation across all runs in the myoelectric
data set. The different transfer learning approaches are listed on the x axis,
the number of data points $\tardatalim$ for transfer learning on the y axis. The best results in
each row are highlighted via bold print.}
\label{tab:myo_results}
\end{table}

In our main analysis, we first consider the case where data from all classes is available for
transfer learning. Again, we iterate over all $236$ runs and treat the data in the current run
as test data, both for the source as well as for the target space. As training data in the source
space, we use the data from all remaining runs of the same subject. We train a slGMM with GMLVQ
initialization and $\protolim = 2$ starting from $5$ random initializations and select the one with
the lowest training error. Then, we use a small random sample of the data from all remaining runs
of the same subject in the alternative electrode configuration as training data for transfer
learning and record the classification error on the (unseen) target space data from the current run.

The mean classification error across all $236$ runs is shown in \tab~\ref{tab:myo_results} and
\fig~\ref{fig:myo_data_results} (left and middle).
Several significant effects can be observed using a one-sided Wilcoxon signed rank test:
\begin{enumerate}
\item Classification performance degrades if an electrode shift is applied, i.e.\ the naive error is
significantly higher than the source error ($p < 10^{-3}$).
\item If at least $12$ data points are available for training,
our proposed algorithm outperforms a naive application of the source space model ($p < 10^{-3}$).
\item If between $12$ and $32$ data points are available, the proposed scheme outperforms a
retrained model on the target data ($p < 10^{-3}$), lending support for H1.
\item If at least $12$ data points are available for training, our proposed algorithm outperforms
the adaptive SVM ($p < 10^{-3}$).
\item If at least $16$ data points are available for training, our proposed algorithm outperforms
gradient-based learning on the GMLVQ cost function ($p < 10^{-3}$).
\end{enumerate}

With regards to runtime, we note that our proposed algorithm is roughly 30 times faster compared to
GMLVQ and a-SVM and roughly 100 times faster compared to re-training a new model on the target
space data (see \fig~\ref{fig:myo_data_results}, right), supporting H3.

\begin{figure}
\begin{center}
\begin{tikzpicture}
\begin{groupplot}[
	group style={
		group size=3 by 2,
		horizontal sep=0.1cm,
		vertical sep=1cm,
		x descriptions at=edge bottom,
		y descriptions at=edge left,
	},
	height=3.47cm,
	width=4.3cm,
	xlabel={$\tardatalim$},
	xmode=log,
	log basis x={2},
	xmin=4,
	xmax=128,
	xtick={4,8,16,32,64,128},
	log basis ticks=2,
	enlarge x limits=0.1,
	enlarge y limits=true,
	ymin=0,
	ymax=0.65,
	ytick={0,0.2,0.4, 0.6},
	legend cell align=left,
	legend pos=outer north east,
	every axis title/.append style={anchor=base}
]


\nextgroupplot[
	title={flexion excl.}
]
\addplot[aluminium3, sharp plot, ultra thin, update limits=false, forget plot]
coordinates {(0.001,0) (1000,0)};
\addplot[em] table[x=n, y=err_mean_vq, y error=err_std_vq] {myo_data_flexion_excluded_errors.csv};
\addplot[retrain] table[x=n, y=err_mean_retrain, y error=err_std_retrain] {myo_data_flexion_excluded_errors.csv};
\addplot[gmlvq] table[x=n, y=err_mean_gmlvq, y error=err_std_gmlvq] {myo_data_flexion_excluded_errors.csv};
\addplot[asvm] table[x=n, y=err_mean_asvm, y error=err_std_asvm] {myo_data_flexion_excluded_errors.csv};

\nextgroupplot[
	title={pronation excl.}
]
\addplot[aluminium3, sharp plot, ultra thin, update limits=false, forget plot]
coordinates {(0.001,0) (1000,0)};
\addplot[em] table[x=n, y=err_mean_vq, y error=err_std_vq] {myo_data_pronation_excluded_errors.csv};
\addplot[retrain] table[x=n, y=err_mean_retrain, y error=err_std_retrain] {myo_data_pronation_excluded_errors.csv};
\addplot[gmlvq] table[x=n, y=err_mean_gmlvq, y error=err_std_gmlvq] {myo_data_pronation_excluded_errors.csv};
\addplot[asvm] table[x=n, y=err_mean_asvm, y error=err_std_asvm] {myo_data_pronation_excluded_errors.csv};

\nextgroupplot[
	title={spread excl.}
]
\addplot[aluminium3, sharp plot, ultra thin, update limits=false, forget plot]
coordinates {(0.001,0) (1000,0)};
\addplot[em] table[x=n, y=err_mean_vq, y error=err_std_vq] {myo_data_spread_excluded_errors.csv};
\addlegendentry{EM (prop.)}
\addplot[retrain] table[x=n, y=err_mean_retrain, y error=err_std_retrain] {myo_data_spread_excluded_errors.csv};
\addlegendentry{retrain}
\addplot[gmlvq] table[x=n, y=err_mean_gmlvq, y error=err_std_gmlvq] {myo_data_spread_excluded_errors.csv};
\addlegendentry{GMLVQ}
\addplot[asvm] table[x=n, y=err_mean_asvm, y error=err_std_asvm] {myo_data_spread_excluded_errors.csv};
\addlegendentry{a-SVM}

\nextgroupplot[
	title={extension excl.}
]
\addplot[aluminium3, sharp plot, ultra thin, update limits=false, forget plot]
coordinates {(0.001,0) (1000,0)};
\addplot[em] table[x=n, y=err_mean_vq, y error=err_std_vq] {myo_data_extension_excluded_errors.csv};
\addplot[retrain] table[x=n, y=err_mean_retrain, y error=err_std_retrain] {myo_data_extension_excluded_errors.csv};
\addplot[gmlvq] table[x=n, y=err_mean_gmlvq, y error=err_std_gmlvq] {myo_data_extension_excluded_errors.csv};
\addplot[asvm] table[x=n, y=err_mean_asvm, y error=err_std_asvm] {myo_data_extension_excluded_errors.csv};

\nextgroupplot[
	title={supination excl.}
]
\addplot[aluminium3, sharp plot, ultra thin, update limits=false, forget plot]
coordinates {(0.001,0) (1000,0)};
\addplot[em] table[x=n, y=err_mean_vq, y error=err_std_vq] {myo_data_supination_excluded_errors.csv};
\addplot[retrain] table[x=n, y=err_mean_retrain, y error=err_std_retrain] {myo_data_supination_excluded_errors.csv};
\addplot[gmlvq] table[x=n, y=err_mean_gmlvq, y error=err_std_gmlvq] {myo_data_supination_excluded_errors.csv};
\addplot[asvm] table[x=n, y=err_mean_asvm, y error=err_std_asvm] {myo_data_supination_excluded_errors.csv};

\nextgroupplot[
	title={fist excl.}
]
\addplot[aluminium3, sharp plot, ultra thin, update limits=false, forget plot]
coordinates {(0.001,0) (1000,0)};
\addplot[em] table[x=n, y=err_mean_vq, y error=err_std_vq] {myo_data_fist_excluded_errors.csv};
\addplot[retrain] table[x=n, y=err_mean_retrain, y error=err_std_retrain] {myo_data_fist_excluded_errors.csv};
\addplot[gmlvq] table[x=n, y=err_mean_gmlvq, y error=err_std_gmlvq] {myo_data_fist_excluded_errors.csv};
\addplot[asvm] table[x=n, y=err_mean_asvm, y error=err_std_asvm] {myo_data_fist_excluded_errors.csv};
\end{groupplot}
\end{tikzpicture}
\end{center}
\caption{Mean classification error across all runs in the myoelectric data set if one movement
was excluded from the training data for transfer learning. The excluded class is listed in the
title of each plot.
The $x$-axis indicates the number of available target space training data points $\tardatalim$ (in log scaling)
while the $y$-axis displays the mean classification error. Error bars indicate the standard deviation.}
\label{fig:myo_data_results_missing}
\end{figure}
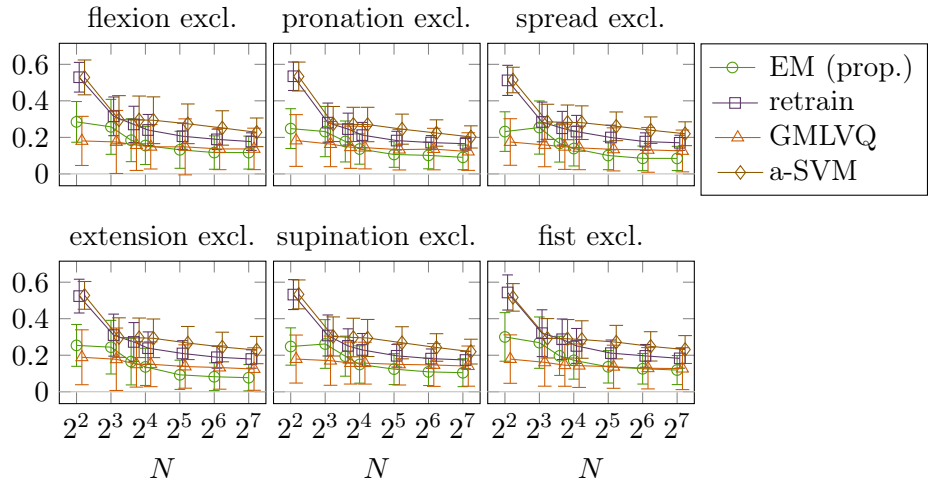

\begin{table}
\begin{center}
\footnotesize
\begin{tabular}{lWWWWW}
$\tardatalim$ & naive           & EM (prop.)              & retrain            & GMLVQ                   & a-SVM \\
\cmidrule(r){1-1} \cmidrule(r){2-2} \cmidrule(r){3-3} \cmidrule(r){4-4} \cmidrule(r){5-5} \cmidrule{6-6}
$4$   & $\bm{18.8 \pm 12.8 \%}$ & $25.4 \pm 11.5 \%$      & $52.4 \pm 9.2 \%$  & $18.9 \pm 15.0 \%$      & $52.8 \pm 7.6 \%$ \\
$8$   & $18.8 \pm 12.8 \%$      & $24.4 \pm 14.7 \%$      & $31.0 \pm 11.5 \%$ & $\bm{17.8 \pm 17.1 \%}$ & $29.7 \pm 10.8 \%$ \\
$12$  & $18.8 \pm 12.8 \%$      & $16.5 \pm 12.8 \%$      & $27.3 \pm 10.6 \%$ & $\bm{15.0 \pm 12.5 \%}$ & $29.5 \pm 10.8 \%$ \\
$16$  & $18.8 \pm 12.8 \%$      & $\bm{13.6 \pm 10.2 \%}$ & $23.9 \pm 8.9 \%$  & $15.1 \pm 12.3 \%$      & $29.3 \pm 10.6 \%$ \\
$32$  & $18.8 \pm 12.8 \%$      & $\bm{9.3 \pm 8.0 \%}$   & $21.0 \pm 7.0 \%$  & $13.8 \pm 11.9 \%$      & $26.7 \pm 9.1 \%$ \\
$64$  & $18.8 \pm 12.8 \%$      & $\bm{8.2 \pm 7.6 \%}$   & $18.9 \pm 6.0 \%$  & $13.3 \pm 11.8 \%$      & $24.5 \pm 8.1 \%$ \\
$128$ & $18.8 \pm 12.8 \%$      & $\bm{7.7 \pm 7.2 \%}$   & $17.9 \pm 5.0 \%$  & $12.4 \pm 11.6 \%$      & $22.8 \pm 7.5 \%$ \\
\end{tabular}
\end{center}
\caption{Mean classification test error and standard deviation across all runs in the myoelectric
data set when no samples for the extension movement were available for transfer learning.
The different transfer learning approaches are listed on the x axis,
the number of data points $\tardatalim$ for transfer learning on the y axis. The best results in
each row are highlighted via bold print.}
\label{tab:myo_results_missing}
\end{table}

With regards to H2, we consider the case of single motions missing from the training data
for transfer learning. This is motivated by the practical application scenario of bionic prostheses.
In practice, as soon as a user notes deteriorating classification accuracy of her hand prosthesis,
she would have to record new labeled training data to learn a transfer mapping which enhances
accuracy again. In this recording process, the user would have to execute a precisely timed
calibration sequence of neural patterns which correspond to desired hand motions.
Any mistakes in timing introduce label noise into the training data and thus may lead to
deteriorating performance. Therefore, any motion which has \emph{not} to be recorded reduces the
likelihood of label noise and enhances ease-of-use.

We repeated our experiments with each of the six motions being missing from the training data
for transfer learning (the resting class can be considered easy to record because users do not have
to actively produce any specific neural pattern). We also experimented with omitting more than one
class in the training data, but observed that under these conditions no transfer method outperformed
the baseline of naively applying the source model to the target space data.

The average results across participants and trials are depicted in
\tab~\ref{tab:myo_results_missing} and \fig~\ref{fig:myo_data_results_missing}.
Several significant effects can be observed using a one-sided Wilcoxon signed rank test:
\begin{enumerate}
\item If at least $32$ data points are available for training,
our proposed algorithm outperforms a naive application of the source space model ($p < 10^{-3}$).
\item Irrespective of the number of available data points, our EM transfer learning scheme
outperforms a retrained model on the target data ($p < 10^{-3}$).
\item If at least $12$ data points are available for training, our proposed algorithm outperforms
the adaptive SVM ($p < 10^{-3}$).
\item If extension, pronation, supination, or spread are excluded and at $32$ data points are
available for training, our proposed algorithm outperforms gradient-based learning on the GMLVQ cost
function ($p < 0.01$).
\end{enumerate}
In conjunction, these results support H2.

\section{Conclusion}

We have presented a novel expectation maximization (EM) algorithm which learns a linear transfer function between
a target and a source space. This transfer function maximizes the likelihood of target space data according to a labeled
Gaussian mixture model in the source space. We demonstrated how this algorithm can be combined
with models from the learning vector quantization family, in particular generalized learning
vector quantization (GMLVQ) and local GMLVQ. We have argued that learning a transfer function
between target and source space is easier compared to re-learning a full classification model
in the target space, if the change in representation between both spaces is structurally simple, that is, approximately
linear.

In our experiments, we evaluated our approach on two artificial data sets and a data set of real electromyographic (EMG)
data from the domain of bionic hand prostheses. In all cases, our proposed method was able to learn a
transfer mapping which significantly improved the classification accuracy compared to all baselines,
in particular if few target space samples were available and a class
was not represented in the training set. The latter aspect is particularly relevant
in settings where recording data for an additional class may be demanding, as is the
case in bionic hand prostheses.

Overall, our proposed transfer learning approach appears as a simple, data- and
time-saving alternative compared to re-learning a new classification model, and even
other domain adaptation and transfer learning approaches. For future work,
it may be interesting to explore whether the learned transfer function can be used
to transfer other classification models than those from the learning vector
quantization family, whether the transfer function can be adjusted on-line,
and whether theoretic guarantees for transfer learning are possible.
Beyond these machine learning questions we hope to use transfer learning
to get one step closer towards an intuitive, rapid, and robust user
interface for bionic prostheses that helps amputees to achieve better
hand function in everyday tasks.

\section*{Acknowledgement}

Funding by the DFG under grant number HA 2719/6-2, the CITEC center of 
excellence (EXC 277), and the EU-Project \enquote{Input} (grant number 687795)
is gratefully acknowledged. We also thank our reviewers for their insightful
comments which helped to improve the quality of our contribution.

\bibliographystyle{abbrvurl}
\bibliography{literature.bib}

\end{document}